\documentclass{article}

\usepackage{microtype}
\usepackage{graphicx}
\usepackage{subcaption}
\usepackage{booktabs}
\usepackage{multirow}
\usepackage{longtable}
\usepackage{fancyvrb}
\usepackage{hyperref}

\usepackage[preprint]{icml2027}

\usepackage{amsmath}
\usepackage{amssymb}
\usepackage[capitalize,noabbrev]{cleveref}
\usepackage{xcolor}

\newcommand{\rec}{\mathrm{rec}}
\newcommand{\fault}[1]{\textsc{#1}}

\icmltitlerunning{Blind Diagnosis and Repair of Deployment Faults in Vision Models}

\begin{document}

\twocolumn[
  \icmltitle{Where Does a Weight-Editing Agent Help? \\
    Blind Diagnosis and Repair of Deployment Faults in Vision Models}

  \icmlsetsymbol{equal}{*}

  \begin{icmlauthorlist}
    \icmlauthor{Zhuangzi Li}{aff}
    \icmlauthor{Shilv Cai}{aff}
    \icmlauthor{Jian Jin}{aff}
 \icmlauthor{Weisi Lin}{aff}
  \end{icmlauthorlist}

  \icmlaffiliation{aff}{Nanyang Technological University, Singapore}

\icmlcorrespondingauthor{Shilv Cai}{shilyu.cai@ntu.edu.sg}

  \icmlkeywords{weight-space programming, model repair, LLM agents, deployment faults, test-time adaptation}

  \vskip 0.3in
]

\printAffiliationsAndNotice{}

\begin{abstract}
Language-model agents can read and edit the weights of neural networks without gradients, and this is
increasingly presented as a new way of repairing models. We ask what such an agent adds, in a setting with
a known answer. A vision model moved into a new system may be broken by a preprocessing, weight-conversion
or output-convention fault; an agent restricted to forward passes must decide whether a fault exists, name
it, and undo it by editing the weights. Over 183 runs on classification, segmentation and detection, scored
against the fault-free model, we find three things. When the kind of fault is known, the agent is not
needed: a script written afterwards repairs 30 of 48 faults against the agent's 33. But the agent did not
have to be told: on five fault types outside every script it repairs 15 of 15 where that script repairs 4,
and it never edits any of 21 fault-free models. And its limits are predictable: it needs a trustworthy
per-layer reference for the model's activations and a fault that is an invertible structured
transformation; without them, small faults are missed and repairs stay partial. We derive guidance on when
to use an agent and when a script.
\end{abstract}

\section{Introduction}
\label{sec:intro}

\begin{figure*}[t]
  \centering
  \includegraphics[width=\textwidth]{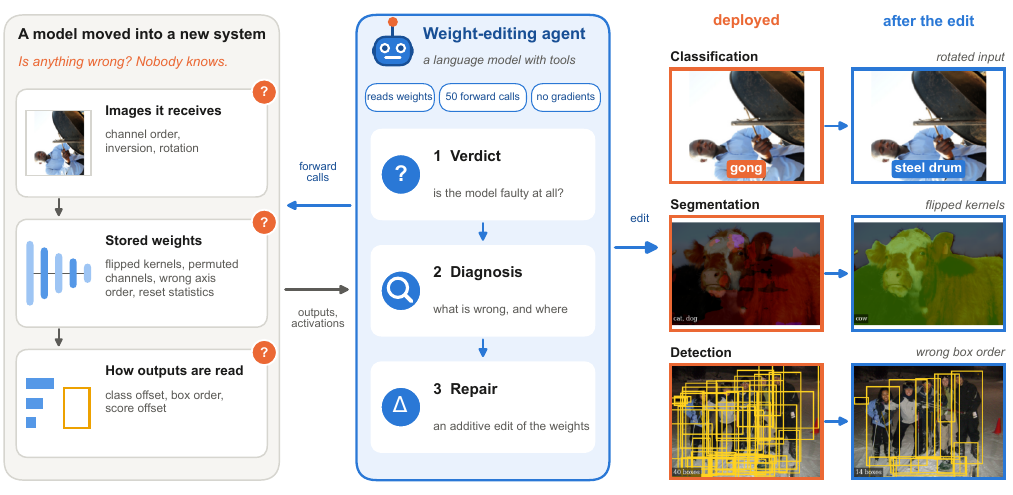}
  \caption{The setting. A vision model has been moved into a new system and may carry a fault in the images
  it receives, in its stored weights, or in the way its outputs are read. An agent that can read the weights
  and run a limited number of forward passes gives a verdict, a diagnosis and an additive edit of the
  weights. Right: outputs of three deployed models before and after the agent's edit (a rotated input in
  classification, flipped kernels in segmentation, a wrong box-coordinate order in detection; examples from
  \cref{fig:qual}).}
  \label{fig:teaser}
\end{figure*}

A trained vision model rarely fails in deployment because its weights are bad. It fails because something
around or inside it was changed on the way: the images arrive with the colour channels in another order or
without normalisation, a converter stored a weight tensor with its axes in the wrong order or flipped the
kernels between the convolution and correlation conventions, the running statistics of a normalisation layer
were reset, or the consumer of the outputs reads box coordinates in another order than the model writes them.
Such faults are silent. The model still runs and returns plausible outputs, and the loss ranges from a few
points to everything. In practice they are found by hand, or they are papered over by generic remedies such as
re-estimating normalisation statistics, test-time adaptation or fine-tuning on a little labelled data, which
treat every fault as a distribution shift.

A different remedy has appeared recently. General-purpose language models, used as agents, can inspect the
weights and activations of a network and write targeted changes into the weights without gradient descent, a
practice named weight-space programming \cite{takehi2026wsp}. Demonstrations include fixing a control policy
stuck in a local optimum and removing a reasoning loop from a small language model \cite{takehi2026wsp},
repairing a planted corruption in a language model \cite{selfrewrite2026}, editing the behaviour of image
classifiers on request \cite{ropinski2026ert}, and writing small transformers from scratch
\cite{mehta2026iwsp}. These studies establish that the capability exists, and several present it as the beginning of a change
in how models are optimised and repaired. They leave open the question a practitioner would ask first:
\emph{what does the agent add over a script?} None of them includes a non-agent method built for the same
job. If the kind of fault is known, a script computes the inverse; an agent is of
interest only if it finds out what is wrong when nobody knows, and if it does so without damaging a model
that was fine.

We study this question for deployment faults of vision models. The agent receives the weights, a small pool
of deployment images and a tool that runs forward passes under a budget. It is not told whether anything is
wrong (except in the first classification round, where it is told that a fault exists). It must hand in a verdict, a diagnosis and an additive edit of the weights. Because every fault we
plant has a known exact or near-exact inverse, the edit can be scored against the fault-free model on a locked
test split, and fault-free models measure false alarms and harm. The comparison is made against controls that
were fixed before any agent ran: a library of forward-only repair scripts and its best combination, entropy
minimisation, and fine-tuning on the labelled pool.

\begin{figure*}[t]
  \centering
  \includegraphics[width=\textwidth]{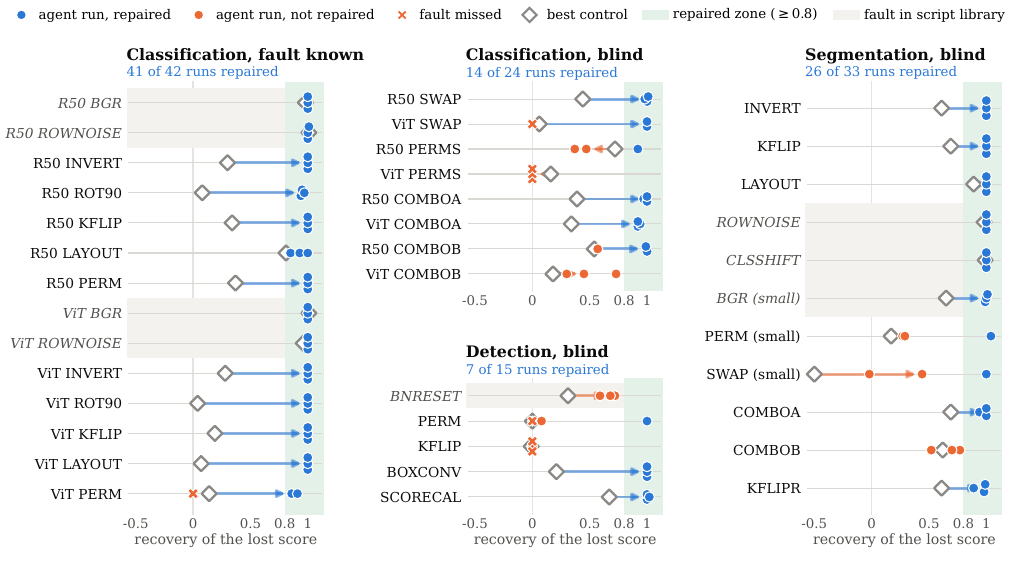}
  \caption{Recovery of every agent run on a faulted model (three seeds per condition) against the best
  non-informed control (largest median among try-all, entropy minimisation and fine-tuning; diamonds). The
  arrow runs from that control to the agent's median. Recovery 1 restores the fault-free score; the shaded
  band marks repaired runs; controls below $-0.5$ are drawn at the left edge. Rows on a grey ground are fault
  types covered by the script library, where agent and control coincide. Outside the library the agent is
  far ahead on the classifiers and the segmentation model, with exceptions for permutations and compound
  faults; on the detector only the two output-convention faults are repaired. Crosses are faulted models
  that the agent declared fault-free.}
  \label{fig:overview}
\end{figure*}

Our findings, over 183 runs of a Claude Opus 5.5 agent on four models and three tasks, cut both ways:
\begin{itemize}
  \item \textbf{What the agent does not add.} Where a frozen script already covers the fault, the agent
    matches it and does not exceed it. More than that: a script written \emph{after} the agent rounds, with
    knowledge of the fault families and of the agents' own solutions, repairs 30 of 48 faulted segmentation
    and detection models against the agent's 33 and also leaves fault-free models alone (\cref{sec:lh}). The
    diagnosis does not need an agent in principle.
  \item \textbf{What it does add.} The agent was never told the fault families. On five fault types that
    are in no script library and that it had not met, it repairs 15 of 15 while the post-hoc script repairs
    4 (\cref{sec:unseen}). Over all rounds it repairs 123 of 162 faulted models, edits none of the 21
    fault-free ones, and no edit lowers the score by more than 0.03 points, whereas fine-tuning and statistics re-estimation often
    do. A correct diagnosis is followed by a repair in 115 of 124 runs, a wrong one in 0 of 13.
  \item \textbf{Where it fails.} (i) The inverse is an assignment problem (a random channel permutation is
    located but its order is not reliably recovered). (ii) A second fault hides behind the first. (iii) The
    model carries no trustworthy per-layer reference for its own activations: on a Vision Transformer faults
    that cost one to three points are declared absent, and supplying reference statistics removes these
    misses; on a detector with frozen, mismatched normalisation statistics the same happens in a
    convolutional backbone.
  \item \textbf{When to use it.} These results give a rule of use (\cref{sec:when}): a script when the fault
    family is known, an agent when it is not and the model offers a reference, and neither as a substitute
    for comparing with the original model when that is available.
\end{itemize}
The results do not rest on remembered values of a public checkpoint: on a privately fine-tuned model
with permuted channels it repairs 8 of 9 faults (\cref{sec:robust}).
\Cref{fig:overview} shows the runs of the main rounds against the best control. We report the expectations we wrote before
each round together with the outcomes, including those that turned out wrong (\cref{app:expect}).

\section{Related Work}
\label{sec:related}

\subsection{Model editing}

\paragraph{Editing on request.}
Model editing changes the behaviour of a trained network on a stated target while leaving the rest intact.
In language models, hypernetworks map a fine-tuning gradient to an edit \cite{decao2021ke,mitchell2022mend},
closed-form low-rank updates are written into located layers \cite{meng2022rome,meng2023memit}, and
memory-based methods route edited inputs around the weights \cite{mitchell2022serac}; \citet{yao2023editing}
survey the field and its failure modes. In vision, a prediction rule of a classifier
\cite{santurkar2021editing} or of a generative model \cite{bau2020rewriting} is rewritten at a chosen layer,
concepts are erased from diffusion models \cite{gandikota2023erasing}, task vectors are added and
subtracted \cite{ilharco2023task}, a concept direction is projected out \cite{ravfogel2020inlp}, and
retraining the last layer removes a spurious correlation \cite{kirichenko2023dfr}. Networks can also be
trained to be editable \cite{sinitsin2020enn} or debuggable \cite{wong2021sparse}. In all of these the
request is given: someone has already decided what is wrong and what the output should be. We study the
step before the edit, in which nobody knows whether anything is wrong.

\paragraph{Adaptation without a request.}
The remedies that are applied in practice when a deployed model underperforms need no diagnosis.
Re-estimating normalisation statistics on deployment data \cite{schneider2020bnadapt,nado2020bn}, entropy
minimisation \cite{wang2021tent}, self-supervised test-time training \cite{sun2020ttt} and alignment of
activation statistics to stored training statistics \cite{mirza2023actmad} adapt the weights to the incoming
data, and fine-tuning on a few labels does the same with supervision. These methods treat every loss of
accuracy as a distribution shift. Deployment faults are of a different kind: a mismatch in resizing or
colour handling between libraries changes results silently \cite{parmar2022aliased}, and conversion
errors change the function the weights compute. We use these adaptation methods as controls and find that
they recover little on structural faults and lower the score of intact models.

\subsection{Weight-space programming}

\paragraph{Weights as data.}
A line of work treats the weights of trained networks as an object of study: accuracy and hyperparameters
are predicted from weights \cite{unterthiner2020predicting}, representations are learned on populations of
models \cite{schurholt2021hyper}, and architectures for weight inputs respect the permutation symmetry of
hidden units \cite{navon2023equivariant,zhou2023neural}. The same symmetry underlies the matching of units
between networks for model merging \cite{ainsworth2023rebasin}; one of our fault types asks for exactly this
matching, and it is where the agents are least reliable. Interpretability gives units a meaning, by
dissection \cite{bau2017dissection} or by language-model explanations \cite{bills2023neurons}, and agents
equipped with interpretability tools run experiments on vision models \cite{shaham2024maia}. Language
models have also been trained to read weights \cite{kabra2026weightoracles}.

\paragraph{Language models that read and write weights.}
\citet{takehi2026wsp} show that a general-purpose frontier model, without such training, names what a small
network was trained for from its raw weights, repairs a control policy stuck in a local optimum, writes
game-playing networks from scratch, and removes a looping behaviour from a 0.8B-parameter language model
within 50 forward passes; they call this weight-space programming and report that editing removed an
obstacle but could not add an ability the model lacked. Follow-up work has moved quickly.
\citet{selfrewrite2026} let language models edit their own weights and include the repair of one planted
corruption, noise written over rows of one matrix. \citet{mehta2026iwsp} let agents write the weights of
small transformers and of brain encoding models under an interpretability constraint.
\citet{ropinski2026ert} benchmark agents that edit image classifiers on request (relabel a concept, fix a
confusion, fix a subgroup) and propose a training method that makes networks easier to edit; the agents'
edits are almost always rank-one changes of the final layer, and accuracy after an edit stays three to
four points below the unedited network. On the reading side, \citet{ahmed2026audit} finds that zero-shot
weight reading by the model we also use combines correct local arithmetic with priors over familiar public
datasets, and fails when the label order of a familiar dataset is permuted. Agents are also used one level
up, to run machine-learning experiments \cite{huang2024mlagentbench}.

\paragraph{Position of this work.}
These studies establish that agents can edit weights, mostly in settings where the target of the edit is
given and success is measured on that target. None of them asks whether the same result
could be had without an agent. One data point suggests the question matters: the corruption repaired in
\citet{selfrewrite2026}, noise written over a few rows of one matrix, is one of our fault types, and in
our vision models a script that zeroes rows with outlying norms recovers 0.94 to 1.00 of the lost score,
with nothing left for the agent to add (\cref{sec:cls}). The models and modality differ, so this does not
say their result is reproduced by a script; it says the comparison should be run. Our setting differs from
the earlier ones in three respects. The agent is not told
that a fault exists or of what kind, and fault-free models measure false alarms and harm. The faults are
deployment and conversion errors with a known inverse, most of them inside the backbone, so that every
edit is scored against the fault-free model instead of against the request. And the comparison is with
scripts, test-time adaptation and fine-tuning that were fixed before any agent ran, and with a script
written afterwards as an upper bound on what a script could have done, so that the question is not whether
an agent can edit weights but what it adds. The reliance on priors reported by
\citet{ahmed2026audit} applies to us as well, since all our main models are public checkpoints; we return
to it in \cref{sec:limits}.

\section{Problem and Protocol}
\label{sec:setup}

\subsection{Task and metric}
A released model $f_\theta$ reaches a reference score $s_\mathrm{ref}$ on its deployment distribution. A
faulted deployment is a pair $(T,\theta')$: the images pass through a fixed pipeline $T$ that cannot be
changed, and the stored weights are $\theta'$. Only the weights may be edited. A method returns an additive
edit $\Delta$ on any entries of the state dictionary, and the deployed model is $f_{\theta'+\Delta}\circ T$.
With $s_\mathrm{def}$ the score of the faulted deployment and $s_\mathrm{edit}$ the score with the edit, both
on a locked test split,
\begin{equation}
  \rec = \frac{s_\mathrm{edit}-s_\mathrm{def}}{s_\mathrm{ref}-s_\mathrm{def}} .
\end{equation}
A run is \emph{repaired} if $\rec\ge0.8$. An edit does \emph{harm} if it lowers the score by more than half a
point. A \emph{miss} is a faulted model declared fault-free; a \emph{false alarm} is a fault-free model that
is edited. For every fault type we build an \emph{informed} edit from the ground truth; for the types we use its
recovery is 1.00 within 0.01, except for one type on one model (0.93), so an edit that restores the model always exists and what is tested
is whether it is found.

\subsection{Models, data and splits}
\Cref{tab:setup} lists the four public checkpoints: ResNet-50 \cite{he2016resnet} and ViT-B/16
\cite{dosovitskiy2021vit}, evaluated on ImageNetV2 \cite{recht2019imagenetv2} in place of the ImageNet
validation set \cite{deng2009imagenet}; DeepLabV3 \cite{chen2017deeplabv3} on PASCAL VOC
\cite{everingham2010voc}; and Faster R-CNN \cite{ren2015fasterrcnn} with a feature pyramid
\cite{lin2017fpn} on COCO \cite{lin2014coco}. For each of three seeds the evaluation images are split
into a labelled pool, an unlabelled pool and a locked test split; agents and controls see only the pools.
All reported scores are computed by us on the test split with one scorer per task; nothing an agent reports
is used.

\begin{table}[t]
\caption{Tasks, models and data. Splits are redrawn for each of 3 seeds. The labelled pool has 500 images
in classification and 200 in segmentation and detection. Classification uses ImageNetV2 as a stand-in for
the ImageNet validation set.}
\label{tab:setup}
\centering\small
\setlength{\tabcolsep}{3pt}
\begin{tabular}{@{}lllr@{}}
\toprule
Task & Model & Data & Metric \\
\midrule
Classification & ResNet-50, ViT-B/16 & ImageNetV2 & top-1 \\
Segmentation & DeepLabV3-R50 & VOC 2012 val & mIoU \\
Detection & Faster R-CNN R50-FPN & COCO val2017 & AP \\
\bottomrule
\end{tabular}
\end{table}

\subsection{Faults}
The faults (\cref{tab:faults}; details in \cref{app:faults}) are of three kinds that occur when a model is
moved between systems. \emph{Preprocessing} faults change the image before it reaches the model.
\emph{Conversion} faults change stored weights or buffers. \emph{Output-convention} faults change how
outputs are indexed or scaled. Where a fault has a strength (how many modules, how much noise) it is set by a
fixed rule: the smallest value on a grid that lowers the pool score by 15\% relative, else the largest. A
fault type is used for the main comparison if at least two of its three instances lose 10\% relative and the
informed edit recovers at least 0.8. Later rounds add, by design, faults that cost only a few points,
compound faults, and a fault after which the normalisation statistics were re-estimated.

\begin{table}[t]
\caption{Fault types. H: held out from the script library.}
\label{tab:faults}
\centering\small
\setlength{\tabcolsep}{3pt}
\begin{tabular}{@{}lp{0.72\columnwidth}@{}}
\toprule
\multicolumn{2}{@{}l}{\emph{Preprocessing}} \\
\fault{bgr} & colour channels reversed \\
\fault{invert} (H) & image replaced by $1-x$ \\
\fault{rot90} (H) & image rotated by $90^\circ$ \\
\multicolumn{2}{@{}l}{\emph{Conversion}} \\
\fault{bnreset} & running statistics of 30\% of the BatchNorm layers reset to 0 / 1 \\
\fault{rownoise} & 8 output rows of one tensor overwritten with noise \\
\fault{perm} (H) & input channels of one layer permuted (\fault{perms}: a subset) \\
\fault{kflip} (H) & kernels flipped in both spatial axes, in one or several layers \\
\fault{layout} (H) & weight tensor stored with a wrong axis order \\
\fault{swap} (H) & weights of two adjacent blocks exchanged \\
\multicolumn{2}{@{}l}{\emph{Output convention}} \\
\fault{clsshift} & class outputs shifted by one index \\
\fault{boxconv} (H) & box regression outputs in $(y,x,h,w)$ order \\
\fault{scorecal} (H) & background logit raised by a constant \\
\multicolumn{2}{@{}l}{\emph{Compound and reference removed}} \\
\fault{comboA} & \fault{invert} + \fault{layout} \\
\fault{comboB} & \fault{perm} + \fault{kflip} in one module each \\
\fault{kflipr} & \fault{kflip}, then all BatchNorm statistics re-estimated on deployment data \\
\bottomrule
\end{tabular}
\end{table}

\paragraph{Formal view.}
Write the released model as $f_\theta$ and let $z$ denote the normalised image. A fault is a pair
$(\tau,\phi)$: the model receives $\tau(z)$ instead of $z$ and its stored weights are $\theta'=\phi(\theta)$,
so that the deployment computes
\begin{equation}
  g(z) = f_{\phi(\theta)}\bigl(\tau(z)\bigr), \qquad
  g_\Delta(z) = f_{\phi(\theta)+\Delta}\bigl(\tau(z)\bigr)
  \label{eq:deploy}
\end{equation}
without and with an edit $\Delta$. Output-convention faults reorder or shift rows of the last layer and are
therefore part of $\phi$. We call a fault \emph{weight-invertible} if there is an edit $\Delta^\star$ with
$g_{\Delta^\star}=f_\theta$ on the deployment data; the informed edit is our construction of $\Delta^\star$.
The faults differ in what is needed to compute it.

\emph{Involutions.} Flipped kernels, $W'_{o,c,u,v}=W_{o,c,k-1-u,k-1-v}$, a wrong axis order, exchanged
blocks and the box-coordinate order $(x,y,w,h)\mapsto(y,x,h,w)$ satisfy $\phi\circ\phi=\mathrm{id}$, so
\begin{equation}
  \Delta^\star=\phi(\theta')-\theta' ,
  \label{eq:involution}
\end{equation}
which is fixed as soon as the fault and the affected tensors are named.

\emph{Affine input faults.} If the true image is an affine function of the received one,
$z=a\odot z'+b$ per channel (inversion: $a=-1$, $b=(1-2m)/s$ for normalisation mean $m$ and standard deviation
$s$), the fault is absorbed by the first convolution $W$ and the running mean $\mu$ of the normalisation layer
that follows it:
\begin{equation}
  \tilde W_{o,c,u,v}=a_c\,W_{o,c,u,v},\qquad
  \tilde\mu_o=\mu_o-\sum_{c,u,v}W_{o,c,u,v}\,b_c ,
  \label{eq:fold}
\end{equation}
exactly for a patch embedding and up to the zero-padded border for a padded convolution. A permutation of the
colour channels is the same construction with the input slices of $W$ permuted.

\emph{Permutations.} If the input channels of a layer are permuted, $W'_{o,c}=W_{o,\pi(c)}$, the inverse
exists for every $\pi$ but $\pi\in S_n$ is unknown ($n=256$ in our convolutional instances). It has to be
estimated from a cost $C_{c,c'}$ between the stored input slice $c$ and the upstream channel $c'$,
\begin{equation}
  \hat\pi=\arg\min_{\pi\in S_n}\sum_{c=1}^{n}C_{c,\pi(c)} ,
  \label{eq:assign}
\end{equation}
where $C$ is built from weight statistics of the two layers or from activation covariances. Unlike
\cref{eq:involution,eq:fold}, the result is only as good as the cost: the edit is exact if $C$ identifies all
$n$ channels and is otherwise a partial repair.

\emph{Non-invertible from deployment data.} Reset normalisation statistics are restored exactly only by the
original buffers. Statistics re-estimated on deployment images are an estimate under a different data
distribution, so $\Delta^\star$ exists but is not computable from what the agent has.

\subsection{Agent}
Each case is handled by one Claude Opus 5.5 agent in an isolated folder that contains the weights, the pools
and a tool that is the only permitted way to run the model. The tool returns predictions, on request the
outputs of any module (pooled, as channel statistics, or in full), accepts a candidate edit, and reports the
task metric when all images of a call are labelled. The budget is 50 forward calls of at most 200 images (500 in classification).
Reading weights is free. Gradients, fitting weights to labels by iterative optimisation, reference
checkpoints, other data and the internet are forbidden; closed-form computations and combinatorial solvers on
arrays already obtained are allowed. The agent hands in a verdict (``no fault found'' or ``fault(s)
found''), a diagnosis of at most 300 words, the edit, a script that rebuilds the edit, and notes that
declare any use of labels. In the first classification round the task text said that a fault exists. In all
later rounds it said that there may be none, one or several, and fault-free models were included. The task
texts are given in \cref{app:task}.

\subsection{Controls}
\emph{Script library.} Forward-only scripts, frozen before the agents ran: re-estimation of BatchNorm
statistics, ridge refit and bias correction of the final layer, best input channel order, input
standardisation folded into the first layer, best cyclic class offset, and removal of rows with outlying
norms (\cref{app:scripts}). \emph{Try-all} applies every script and every ordered pair, selects on one half of
the labelled pool and refits the winner on all of it; it stands for an engineer who runs the whole toolbox.
\emph{Entropy minimisation} \cite{wang2021tent} on the unlabelled pool (classification and segmentation).
\emph{Fine-tuning} of the last layer and of all weights on the labelled pool. The \emph{informed} edit is the
upper reference. Fault types are marked as in-library or held out before any agent run.

\subsection{Integrity of the comparison}
For every round the conditions, measures and expectations were written into a log before it ran and only
appended to afterwards. Reference checkpoints and answer keys were moved off the machine's cache while agents
ran. After each round an automatic scan of the agent transcripts for rule violations was reviewed hit by hit, and
the diagnoses were judged against the key, in both cases before the scores of that round were looked at.
Runs in a grey area of the rules are reported with and without. Diagnoses were judged by a single judge, the AI coding assistant that ran the experiments (see
the AI Use Statement), on written criteria: \emph{correct} (mechanism and location), \emph{partly} (location without mechanism, or one
of two components), \emph{wrong}.

\section{Classification}
\label{sec:cls}

\subsection{A fault is known to exist}
With seven fault types on two models, 41 of 42 runs are repaired (\cref{tab:cls1}). On the two types that
the script library covers, the agent is within 0.1 of try-all on both models: it adds nothing. On the five
held-out types the agent's median recovery exceeds try-all by at least 0.3 on four types for ResNet-50 and
on all five for ViT-B/16. Entropy minimisation and fine-tuning stay below the agent on every held-out type,
mostly far below, and are often negative. The agents used between 2 and 36 of the 50 calls. The single failure is a ViT
permutation that the agent did not find. Two of our expectations were wrong in the agent's favour: we had
expected flipped kernels to be found in at most two of six runs and a rotated input to be repairable on
ResNet-50 in at most one of three; both were found and repaired in all runs.

\begin{table}[t]
\caption{Classification, fault known to exist: median recovery over 3 seeds; repaired runs in brackets.}
\label{tab:cls1}
\centering\small
\setlength{\tabcolsep}{3.5pt}
\begin{tabular}{@{}llrrrr@{}}
\toprule
 & Fault & Agent & Try-all & Tent & FT full \\
\midrule
\multirow{7}{*}{\rotatebox{90}{ResNet-50}}
 & \fault{bgr}        & 1.00 (3/3) & 0.98 & 0.18 & 0.18 \\
 & \fault{rownoise}   & 1.00 (3/3) & 1.00 & 1.01 & 0.94 \\
 & \fault{invert} (H) & 1.00 (3/3) & 0.18 & 0.17 & 0.30 \\
 & \fault{rot90} (H)  & 0.95 (3/3) & $-$0.01 & 0.08 & 0.04 \\
 & \fault{kflip} (H)  & 1.00 (3/3) & 0.18 & 0.34 & 0.33 \\
 & \fault{layout} (H) & 0.93 (3/3) & 0.74 & 0.81 & 0.77 \\
 & \fault{perm} (H)   & 1.00 (3/3) & 0.04 & 0.37 & $-$0.20 \\
\midrule
\multirow{7}{*}{\rotatebox{90}{ViT-B/16}}
 & \fault{bgr}        & 1.00 (3/3) & 1.01 & 0.03 & $-$0.61 \\
 & \fault{rownoise}   & 1.00 (3/3) & 0.96 & 0.00 & 0.32 \\
 & \fault{invert} (H) & 1.00 (3/3) & 0.17 & 0.05 & 0.28 \\
 & \fault{rot90} (H)  & 1.00 (3/3) & 0.00 & 0.04 & $-$0.36 \\
 & \fault{kflip} (H)  & 1.00 (3/3) & 0.00 & 0.19 & $-$0.62 \\
 & \fault{layout} (H) & 1.00 (3/3) & $-$0.01 & 0.07 & $-$0.13 \\
 & \fault{perm} (H)   & 0.86 (2/3) & 0.00 & 0.14 & $-$0.36 \\
\bottomrule
\end{tabular}
\end{table}

\subsection{Not told whether a fault exists}
A result near the ceiling does not show where a method stops working. The second round was built to be
harder: the agent is not told that a fault exists, a fifth of the models have none, some faults cost only one
to five points, and some models carry two faults (\cref{tab:cls2}). None of the six fault-free models was
edited, and no run lowered accuracy. Of the 24 faulted models 14 are repaired\footnote{One instance loses
less than one point and is excluded by a rule fixed in advance: 13 of 23.}, 8 of 11 counted runs on
ResNet-50 and 5 of 12 on ViT-B/16. Four faults that cost 1.4 to 2.6 points were declared absent, all four on
ViT-B/16. On compound faults a repair does not mean that both faults were found: on \fault{comboA} undoing
the inversion alone gives a recovery of 0.92 to 0.97, and no ViT run found both components of either
compound. On partial permutations in ResNet-50, entropy minimisation (median 0.72) did better than two of
the three agents. We had expected exchanged blocks to be found in at most three of six runs; five were found
and repaired, from weight statistics and BatchNorm consistency, even where the loss was about one point.

\begin{table}[t]
\caption{Classification, existence of a fault unknown: agent recovery per seed and control medians.}
\label{tab:cls2}
\centering\small
\setlength{\tabcolsep}{3pt}
\begin{tabular}{@{}llcrrr@{}}
\toprule
Fault & Model & Agent (3 seeds) & Try-all & Tent & FT full \\
\midrule
\fault{swap}   & R50 & 1.00, 0.98, 1.01 & 0.00 & 0.44 & $-$2.05 \\
\fault{swap}   & ViT & 0.00, 1.00, 1.00 & $-$0.04 & 0.06 & $-$2.82 \\
\fault{perms}  & R50 & 0.92, 0.37, 0.47 & 0.00 & 0.72 & $-$0.64 \\
\fault{perms}  & ViT & 0.00, 0.00, 0.00 & $-$0.05 & 0.16 & $-$5.05 \\
\fault{comboA} & R50 & 0.97, 1.00, 1.00 & 0.26 & 0.28 & 0.39 \\
\fault{comboA} & ViT & 0.92, 0.94, 0.92 & 0.14 & 0.03 & 0.34 \\
\fault{comboB} & R50 & 1.00, 0.99, 0.57 & 0.34 & 0.54 & 0.40 \\
\fault{comboB} & ViT & 0.73, 0.30, 0.45 & 0.04 & 0.18 & 0.07 \\
\bottomrule
\end{tabular}
\end{table}

\subsection{The role of a per-layer reference}
\label{sec:ref}
ResNet-50 stores, in every BatchNorm layer, the mean and variance its activations had in training. The
ResNet agents used these as a reference: a fault shows as the first layer whose measured statistics leave
the stored ones. In its simplest form, with $v_{\ell,c}$ the stored variance of channel $c$ in layer $\ell$ and
$\hat v_{\ell,c}$ the variance measured on deployment images, the deviation of a layer is
\begin{equation}
  d_\ell=\operatorname*{median}_{c}\,\Bigl|\log\frac{\hat v_{\ell,c}}{v_{\ell,c}}\Bigr| ,
  \label{eq:dev}
\end{equation}
and a fault is located at the first layer in forward order whose $d_\ell$ leaves the range of the layers
before it. ViT-B/16 stores nothing comparable, and all four misses were on it. To test whether the
missing reference is the cause, we repeated the ViT instances of the second round in two arms: one as before,
one in which the agent additionally received per-module activation statistics of the fault-free model on
other images. With the reference 8 of 12 faulted models were repaired and none was missed; without it 4 of
12 were repaired and 4 were missed, as in the first attempt on the same instances (5 of 12 repaired, 4
misses). The reference also halved the median number of calls (6 against 11). It did not help with the second
fault of \fault{comboB}: in neither arm did any run find the flipped patch-embedding kernel.

Does a script do as well when it is given the same reference? We ran three: a closed-form sequential
alignment of every module to the reference, an activation-matching objective minimised by gradient descent,
and a selection among these by the labelled pool. None repaired any of the 12 instances (largest recovery
0.46), and the closed-form alignment lowered the accuracy of all three fault-free models by 1.1 to 1.8
points. A simple detector built on the reference flags all 12 faulted models and one of three fault-free
ones, and points to the right place in 4 of 6 single faults. \emph{Whether} something is wrong is therefore
largely scriptable given a reference, and \emph{where} partly; what the scripts did not reproduce is turning
a located deviation into a specific cause and its exact inverse.

\section{Semantic Segmentation}
\label{sec:seg}
On DeepLabV3 the agent was never told whether a fault exists. In the first round (five valid fault types and
fault-free models) all 15 faulted models were repaired with recoveries of at least 0.995, all diagnoses were
correct, and the three fault-free models were left untouched (\cref{tab:seg}); the agents used 2 to 17 calls.
On the two in-library types the agent again only matches try-all. On the held-out types it is far above
try-all and above full fine-tuning (0.61 to 0.89), which is the strongest control on this model.

The second round was again built to find the limit: three faults that cost one to seven points, two compound
faults, and \fault{kflipr}, in which the kernels of six layers are flipped and all BatchNorm statistics are
then re-estimated on deployment data, so that stored and measured statistics agree everywhere and the
reference used in every earlier diagnosis is gone. Of 17 counted runs 11 were repaired. All 18 models were
recognised as faulty. The failures are of two kinds. A permuted layer was located in all three runs but its
order was recovered in one; the other two switched the damaged residual branch off (recovery 0.27 and 0.29).
In \fault{comboB} the permutation was found and the single flipped kernel behind it was found in none of
three runs, the remaining loss being attributed to the data. Full fine-tuning (0.62) comes close to the
agent's median (0.70) here and exceeds it on one seed.

\fault{kflipr} went against our expectation. We had predicted that without the statistics reference at most
one of three runs would be repaired. All three were diagnosed and repaired: the agents found that the
spatial energy pattern of the kernels in one stage is reversed relative to the other stages, confirmed the
orientation by trials scored on the labelled pool, flipped the kernels back, and re-estimated the statistics
level by level. The repair consumed 47 to 48 of the 50 calls. A structured weight fault can thus be found from
the weights alone when it leaves a signature there.

\begin{table}[t]
\caption{Segmentation (DeepLabV3-R50, VOC): agent recovery per seed and control medians.}
\label{tab:seg}
\centering\small
\setlength{\tabcolsep}{3pt}
\begin{tabular}{@{}lcrrr@{}}
\toprule
Fault & Agent (3 seeds) & Try-all & Tent & FT full \\
\midrule
\multicolumn{5}{@{}l}{\emph{First round}} \\
\fault{invert} (H)   & 1.00, 1.00, 1.00 & 0.29 & $-$0.13 & 0.61 \\
\fault{kflip} (H)    & 1.00, 1.00, 1.00 & 0.32 & 0.12 & 0.69 \\
\fault{layout} (H)   & 1.00, 1.00, 1.00 & 0.76 & 0.74 & 0.89 \\
\fault{rownoise}     & 1.00, 1.00, 1.00 & 0.94 & 0.95 & 0.98 \\
\fault{clsshift}     & 1.00, 1.00, 1.00 & 0.99 & 0.00 & 0.60 \\
\midrule
\multicolumn{5}{@{}l}{\emph{Second round}} \\
\fault{bgr} (small)  & 0.99, 1.00, 1.01 & 0.65 & $-$0.87 & $-$0.61 \\
\fault{perm} (small) & 1.04, 0.27, 0.29 & 0.00 & $-$0.74 & 0.17 \\
\fault{swap} (small) & 0.44, $-$0.02, 1.00 & $-$0.97 & $-$2.05 & $-$0.58 \\
\fault{comboA}       & 0.94, 1.00, 1.00 & 0.38 & 0.11 & 0.69 \\
\fault{comboB}       & 0.52, 0.77, 0.70 & 0.13 & 0.17 & 0.62 \\
\fault{kflipr}       & 0.98, 0.89, 0.99 & 0.20 & $-$0.08 & 0.61 \\
\bottomrule
\end{tabular}
\end{table}

\section{Object Detection}
\label{sec:det}
The detector differs from the other models in one respect that turns out to matter. Its BatchNorm layers are
frozen, and their stored statistics do not describe its own activations: re-estimating them on deployment
data lowers the AP of the fault-free detector from 0.38 to about 0.12. The per-layer reference is present in
name only.

The results change accordingly (\cref{tab:det}). Seven of 15 faulted models were repaired. The two faults in
the output convention of the box head were named and inverted exactly in all six runs, with four to seven
calls. For \fault{scorecal} two of the three agents needed no labels: they observed that the class biases
of a softmax layer trained from zero should sum to about zero, found a sum of 6.03 that sat entirely in the
background entry, and subtracted it. We had expected this fault to be the hardest to name and the agent to
be no more than 0.2 above last-layer fine-tuning (0.67); it reached 1.00 in all three runs.

The argument can be stated exactly. Let the classifier compute logits $u=Wh+\beta$ over $K$ classes and be
trained with the cross-entropy loss $L=-\log p_y$, $p=\mathrm{softmax}(u)$.
\begin{equation}
  \frac{\partial L}{\partial\beta_k}=p_k-\mathbb{1}[k=y]
  \quad\Longrightarrow\quad
  \sum_{k=1}^{K}\frac{\partial L}{\partial\beta_k}=0 .
  \label{eq:biasgrad}
\end{equation}
Hence any optimiser whose update of $\beta$ is linear in the gradient (SGD with or without momentum) keeps
$\mathbf{1}^\top\beta$ at its initial value, and weight decay can only shrink it towards zero. For a layer
initialised near zero, $\mathbf{1}^\top\beta\approx0$ after training, and an offset $s$ added to one entry
is read off the stored weights as
\begin{equation}
  \hat s=\mathbf{1}^\top\beta' .
  \label{eq:shat}
\end{equation}
In our instances $\hat s=6.03$ for a true offset of $6$, and the same identity holds for the sum over classes of
the rows of $W$, which the agents used to check that only the bias had been changed. The diagnosis thus rests on an invariant
of training that is visible in the weights, not on a reference copy of them.

Inside the backbone the agent largely failed. Flipped kernels, repaired as a single fault in every run on the classifiers and
on the segmentation model, were repaired in none of three. In the two instances with one flipped layer
(losses of 5 and 9 AP points) the agents measured a pool AP that was, by their own bootstrap, within noise
of the reported figure, found no break in the statistics, and declared the model fault-free after five and
three calls. In the instance with six flipped layers (AP 0.08) the agent recognised a fault, spent 36 calls,
did not localise it, and asserted and edited a fault that does not exist. Of the three permutations one was
recovered exactly, one was located but only ablated, and one was missed. Reset statistics were diagnosed in
all three runs and re-estimated layer by layer; recovery reached 0.59 to 0.72, twice that of try-all, but
not 0.8. The agents stated the reason themselves: the original statistics came from training data, and
deployment data only approximate them. This is a fault that is diagnosed with certainty and whose inverse
needs information the deployment data do not contain.

No fault-free detector was edited and no edit lowered AP, in a setting where full fine-tuning lowers it for
three of the five fault types.

\begin{table}[t]
\caption{Detection (Faster R-CNN R50-FPN, COCO): agent recovery per seed and control medians.}
\label{tab:det}
\centering\small
\setlength{\tabcolsep}{3pt}
\begin{tabular}{@{}lcrrr@{}}
\toprule
Fault & Agent (3 seeds) & Try-all & FT last & FT full \\
\midrule
\fault{bnreset}      & 0.72, 0.59, 0.68 & 0.31 & 0.00 & 0.02 \\
\fault{perm} (H)     & 0.08, 1.00, 0.00 & 0.00 & $-$0.51 & $-$1.82 \\
\fault{kflip} (H)    & 0.00, 0.00, 0.00 & $-$0.01 & $-$0.11 & $-$0.28 \\
\fault{boxconv} (H)  & 1.00, 1.00, 1.00 & 0.00 & 0.09 & 0.21 \\
\fault{scorecal} (H) & 1.00, 1.00, 1.02 & 0.00 & 0.67 & $-$0.09 \\
\bottomrule
\end{tabular}
\end{table}

\begin{figure*}[t]
  \centering
  \includegraphics[width=0.93\textwidth]{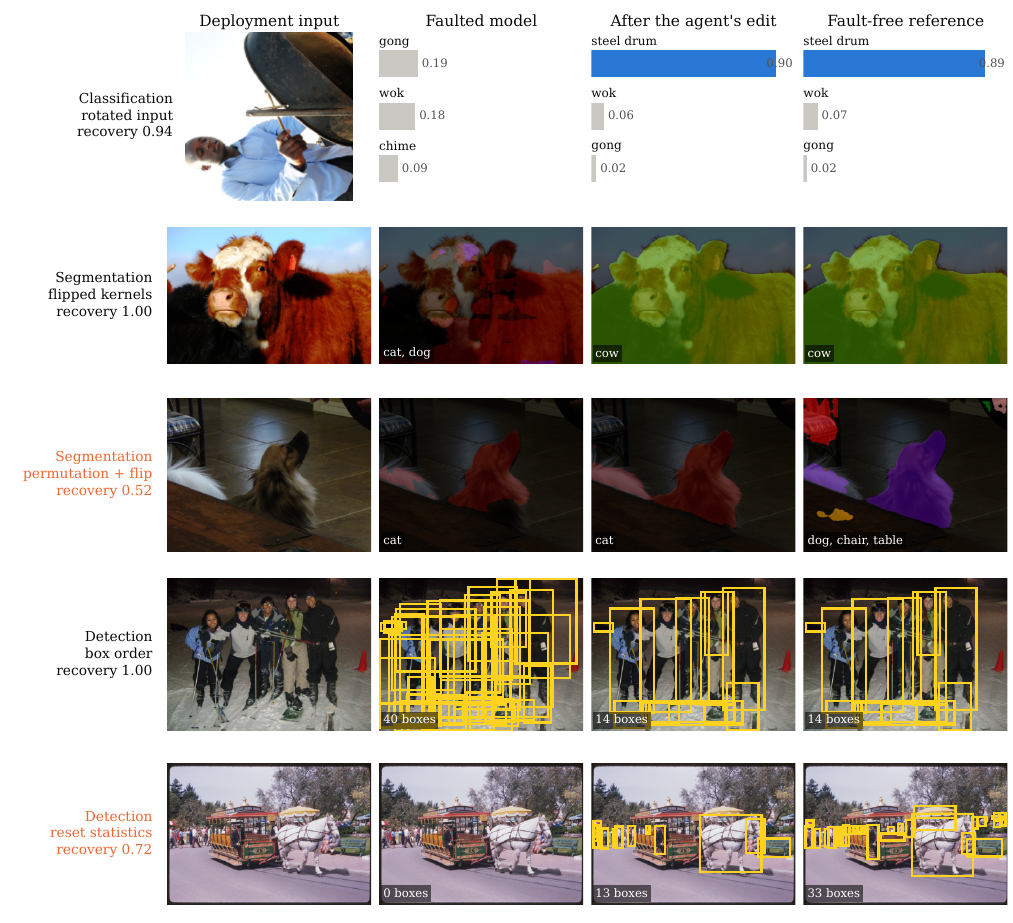}
  \caption{Outputs of the deployed model before and after the agent's edit, with the fault-free model for
  reference; one agent run per row, labelled pool images only. Classification: top-3 classes (blue: the
  correct class). Segmentation: predicted label map over the image. Detection: boxes with score at least
  0.5. Rows in orange were not repaired (recovery below 0.8). The image of a row is chosen by a fixed rule:
  for repaired runs the pool image, among the first 40, on which the faulted model disagrees most with the
  fault-free model; for runs that were not repaired, the one on which the edited model still disagrees
  most with it, so that the row shows what remains wrong.}
  \label{fig:qual}
\end{figure*}

\section{Is an Agent Needed?}
\label{sec:needed}
The rounds above compare the agent with methods that do not diagnose. They cannot say whether the diagnosis
itself needs an agent. Two further experiments address this from opposite sides.

\subsection{A script written after the fact}
\label{sec:lh}
We wrote the strongest script we could after the six main agent rounds were finished, using everything we had
learned from the fault families and from the agents' solutions. It receives exactly what an agent receives
and runs forward passes only. It first \emph{locates}: one pass over 50 unlabelled images gives the
deviation $d_\ell$ of \cref{eq:dev} for every normalisation layer, and the layers with the largest
deviation, plus the first one above twice the median, become suspects. It then \emph{tests a library of
hypotheses}: input channel orders and inversion; class shifts, box-coordinate orders and the bias-sum
correction of \cref{eq:shat}; for each suspect module kernel flips, transposition, the inverse axis order,
exchange with the next block, switching the residual branch off, and re-estimated statistics; and globally
flipped kernels per stage, with and without re-estimation of all statistics. A hypothesis is accepted if it
raises the score on both halves of the labelled pool by at least one point, and the procedure is repeated
up to three times so that compound faults can be handled (\cref{app:scripts}).

\begin{table}[t]
\caption{Repaired faulted models: the post-hoc script (LH) against the agent on the same instances. LH was
written with knowledge of these fault families; the agent was not told them.}
\label{tab:lh}
\centering\small
\setlength{\tabcolsep}{4pt}
\begin{tabular}{@{}llrr@{}}
\toprule
 & Fault & LH & Agent \\
\midrule
\multirow{7}{*}{\rotatebox{90}{Segmentation}}
 & \fault{invert}, \fault{kflip}, \fault{layout}, & & \\
 & \quad \fault{rownoise}, \fault{clsshift} & 15 / 15 & 15 / 15 \\
 & \fault{bgr}, \fault{perm}, \fault{swap} (small) & 2 / 9 & 5 / 9 \\
 & \fault{comboA} & 3 / 3 & 3 / 3 \\
 & \fault{comboB} & 0 / 3 & 0 / 3 \\
 & \fault{kflipr} & 3 / 3 & 3 / 3 \\
 & Total & 23 / 33 & 26 / 33 \\
\midrule
\multirow{6}{*}{\rotatebox{90}{Detection}}
 & \fault{boxconv}, \fault{scorecal} & 6 / 6 & 6 / 6 \\
 & \fault{kflip} & 1 / 3 & 0 / 3 \\
 & \fault{perm} & 0 / 3 & 1 / 3 \\
 & \fault{bnreset} & 0 / 3 & 0 / 3 \\
 & Total & 7 / 15 & 7 / 15 \\
\midrule
 & Fault-free models edited & 0 / 6 & 0 / 6 \\
\bottomrule
\end{tabular}
\end{table}

\Cref{tab:lh} gives the result. The script repairs 30 of the 48 faulted models, the agent 33, and neither
touches a fault-free model. They agree on most families. The agent is ahead on three small-drop
segmentation faults and one detection permutation, and its partial repair of reset statistics is better
(0.59--0.72 against 0.29--0.33). The script is ahead on the six flipped layers in the detector, which no
agent localised in 36 calls but which the hypothesis ``all kernels of stage 3 flipped'' finds at once. So
for the fault families of this paper the diagnosis can be scripted, and the statement that it needs an
agent is wrong. The comparison is not a fair race and is not meant as one: the script was built from the
families and from the agents' solutions, and bounds from above what a script could have done.

\subsection{Faults nobody prepared for}
\label{sec:unseen}
What is then left of the agent's advantage is that it was not told the families. To test this directly we
built five fault types that are in no script library of this study and that no agent had seen: the sign of
a convolution weight inverted (\fault{sign}); $3\times3$ kernels rolled by one position in both axes, as
when a kernel origin is off by one (\fault{kroll}); the input channels of a convolution in reversed order
(\fault{crev}); three pairs of classes exchanged in the final layer (\fault{pairswap}); and weight and bias
of normalisation layers exchanged (\fault{bnswap}). Each has an exact inverse. The agents, the unchanged
script and the controls were run on three seeds each.

\begin{table}[t]
\caption{Fault types outside every script library (DeepLabV3-R50): recovery per seed of the agent and of the post-hoc script LH, and control medians over 3 seeds. Agent: 15 of 15 repaired; LH: 4 of 15.}
\label{tab:unseen}
\centering\small
\setlength{\tabcolsep}{2.6pt}
\begin{tabular}{@{}lccrrr@{}}
\toprule
Fault & Agent (3 seeds) & LH (3 seeds) & Try-all & Tent & FT full \\
\midrule
\fault{sign} & 1.00, 1.00, 1.00 & 0.96, 0.84, 0.92 & 0.84 & 0.82 & 0.91 \\
\fault{kroll} & 0.97, 1.00, 1.00 & 0.68, 0.70, 0.69 & 0.39 & 0.24 & 0.69 \\
\fault{crev} & 1.00, 1.00, 1.00 & 0.41, 0.00, 0.00 & 0.13 & $-$0.08 & 0.62 \\
\fault{pairswap} & 1.00, 1.00, 1.00 & 0.00, 0.00, 0.00 & 0.69 & $-$0.05 & 0.49 \\
\fault{bnswap} & 1.00, 1.00, 1.00 & 0.61, 0.20, 0.90 & 0.64 & 0.66 & 0.91 \\
\bottomrule
\end{tabular}
\end{table}

The agents repaired all 15 models with a median of four forward calls, and all 15 diagnoses were correct
(\cref{tab:unseen}). They found the inverted sign from an input mean that was the negative of the stored
one with matching variances, the rolled kernels from an energy profile displaced by one tap, the reversed
channels from the lost correspondence with the preceding layer, the exchanged classes by matching the main
head against the auxiliary head, and the exchanged normalisation parameters from weights that were mostly
negative. The script built for the earlier families repaired 4 of 15, by ablation or re-estimation in the
cases where that happens to compensate. We had expected the agents to repair 9 to 12.

Two qualifications. The construction is one-sided: the faults were chosen by us, knowing the script, to
lie outside it, and a script extended by five hypotheses would repair them. The result therefore does not
say that a script cannot do this, only that a script has to be told and the agent did not. And the generic
controls are stronger on two of the five types than on earlier held-out faults: an inverted sign in one
layer is largely compensated by re-estimating the statistics behind it (try-all and fine-tuning above
0.8), so the agent's margin there is small; it is large for rolled kernels, reversed channels and
exchanged classes.

\subsection{Robustness of the findings}
\label{sec:robust}
\paragraph{Non-public weights.}
All main models are public checkpoints, and weight reading by this agent model has been shown to lean on
priors over familiar models \cite{ahmed2026audit}. We therefore repeated three weight-level conditions of
the segmentation rounds on a model that has never been published: the DeepLabV3 checkpoint fine-tuned on the
VOC training split (test mIoU 0.79--0.81 against 0.78), with the internal channels of every bottleneck
permuted at random, which leaves the function unchanged. The agents repaired 8 of 9 faulted models
(\fault{kflip} 3 of 3, \fault{layout} 3 of 3, \fault{kflipr} 2 of 3 with recoveries 0.98, 0.71, 0.98), all
12 diagnoses were correct, and the three fault-free models were not edited; on the public checkpoint the
same conditions gave 9 of 9. The evidence the agents used was again internal to the model. This rules out
that the diagnoses rest on remembered values or positions of the public file. It does not rule out general
knowledge of how such networks look: fine-tuning moved the convolution weights by about 4\% per tensor, and
two agents' notes refer to quirks of public checkpoints.

\paragraph{A second judge.}
The diagnoses were judged by a single judge with the key in view (\cref{sec:setup}). As a check, a different language model
(Claude Opus 5.5 in a fresh context) judged the 156 diagnoses of the six main rounds from the key, the
written criteria and the diagnosis text alone, without scores or the first judgements. The two judges
agree on 154 of 156 labels (Cohen's $\kappa=0.97$); the two disagreements are partial-permutation runs that
the second judge rates wrong instead of partly correct. With the second judge's labels a correct diagnosis
is followed by a repair in 92 of 100 runs, a partly correct one in 8 of 23 and a wrong one in 0 of 15. Both
judges are language models reading explicit criteria, so this shows that the criteria are applied
consistently, not that a human panel would agree.

\section{What Decides Success}
\label{sec:analysis}

\Cref{fig:qual} shows what the numbers mean for the outputs of the deployed models.

\begin{table}[t]
\caption{All agent rounds. Repaired: recovery $\ge0.8$ among faulted models. The reference ablation counts
both arms. Diagnosis judged before scores were seen.}
\label{tab:all}
\centering\small
\setlength{\tabcolsep}{3pt}
\begin{tabular}{@{}lrrrr@{}}
\toprule
Round & Runs & Fault-free & Repaired & Misses \\
 & & (edited) & & \\
\midrule
Classification, fault known & 42 & -- & 41 / 42 & 1 \\
Classification, blind       & 30 & 6 (0) & 14 / 24 & 4 \\
ViT reference ablation      & 30 & 6 (0) & 12 / 24 & 4 \\
Segmentation, first         & 18 & 3 (0) & 15 / 15 & 0 \\
Segmentation, second        & 18 & -- & 11 / 18 & 0 \\
Detection                   & 18 & 3 (0) & 7 / 15 & 3 \\
Segmentation, private weights & 12 & 3 (0) & 8 / 9 & 0 \\
Segmentation, unseen faults & 15 & -- & 15 / 15 & 0 \\
\midrule
Total                       & 183 & 21 (0) & 123 / 162 & 12 \\
\midrule
\multicolumn{5}{@{}l}{Diagnosis correct: 115 / 124 repaired; partly: 8 / 25; wrong: 0 / 13.} \\
\bottomrule
\end{tabular}
\end{table}

\paragraph{Diagnosis decides, and the edit is the easy part.}
\Cref{tab:all} pools the eight rounds. When the diagnosis names the mechanism and the place, the repair
follows in 115 of 124 runs; the nine exceptions are faults whose inverse could not be computed or was only
partly applied (below). When the diagnosis is wrong there is no repair. When the type is told to a script
by the ground truth (the informed edit) it does as well or better. This is also why a script can replace
the agent once the families are known (\cref{sec:lh}): what has to be found is which of a few structural
transformations happened and where, and that can be enumerated.

\paragraph{What the agent adds is not having to enumerate.}
On families a script was written for, script and agent are level (30 against 33 of 48). On families nobody
wrote a script for, the agent keeps its level and the script does not (15 against 4 of 15). The agent
builds the test it needs from the evidence in front of it: an anti-correlated mean suggests a sign, a
displaced energy profile a shift, a lost correspondence a reordering. A script encodes the tests someone
thought of beforehand.

\paragraph{The inverse must be computable from what the agent has.}
Flips, axis orders, index shifts, exchanged blocks and input transforms have inverses that are fixed once
the fault is named, and these were restored to a recovery of about 1.00. A channel permutation needs a
256-way assignment estimated from statistics; over all rounds it was sometimes solved exactly and sometimes
replaced by switching the damaged branch off. Reset statistics in the detector could not be restored from
deployment data at all. The eight repairs that followed a partly correct diagnosis are ablations of a located module or the removal
of the larger of two faults.

\paragraph{A reference for what is normal.}
All twelve misses occurred where the model offers no usable per-layer reference: nine on ViT-B/16 without
reference statistics and three on the detector. No fault was missed on
ResNet-50 or DeepLabV3, including faults that cost about one point, and none on ViT-B/16 once reference
statistics were supplied. Without a reference, a fault that costs a few points has to be seen in the score
on a pool of 200 to 500 images, where it lies within sampling noise; the agents reasoned exactly so and
stopped after a few calls. The detector shows that the reference must be trustworthy, not merely present.

\paragraph{The first fault hides the second.}
In compound cases the agents stopped once the large fault was repaired and the remaining loss looked like
noise or data shift. The single flipped kernel behind a permutation was found by none of the nine runs on
ViT-B/16 (both arms of the ablation and the blind round) and by none of three on DeepLabV3.

\paragraph{No harm.}
In 183 runs no fault-free model was edited and no edit lowered the score by more than the harm threshold.
The agents that could not repair a fault handed in an empty or partial edit and said so. The generic
remedies behave differently: on fault-free models, statistics re-estimation, entropy minimisation and full
fine-tuning cost 0.4 to 7 mIoU points in segmentation, full fine-tuning costs 1.9 to 13 accuracy points in
classification, and re-estimation costs about 26 AP points in detection.

\paragraph{Assertions that do not hold up.}
Three segmentation agents and one detection agent asserted a fault that was not there (a noisy final
classifier, a corrupted variance buffer), from statistics that are in fact properties of the released
weights. Only the detection agent edited it, without effect on AP. Stated doubt was informative in
classification: runs whose diagnosis expressed doubt were repaired less often than those that did not
(6 of 14 against 8 of 10 in the blind round).

\paragraph{Cost.}
In the seven rounds where the agent was not told that a fault exists, the median number of forward calls
was between 4 and 11 of the 50 allowed; the limit was approached only by the re-estimation of all statistics
in \fault{kflipr}.

\section{When to Use an Agent}
\label{sec:when}
The findings translate into a rule of use.

\emph{Compare with the original model first.} Our setting assumes that no reference copy of the weights is
available. Where the original model and a few inputs are at hand, comparing outputs and weights of the two
locates most conversion errors directly, and neither an agent nor our scripts are needed.

\emph{Use a script when the fault family is known.} For channel order, normalisation, class offsets, row
noise, flipped kernels, axis orders, box conventions and score offsets, a script that tests the hypothesis
is as accurate as the agent, runs in seconds to minutes, and is deterministic. The hypothesis library of
\cref{sec:lh} was assembled from the agents' own solutions; an agent run is a way of finding out which
checks are worth adding to it.

\emph{Use an agent when the cause is unknown} and all of the following hold: the model carries a
trustworthy reference for its own activations (normalisation statistics that match them, or supplied
reference statistics); the suspected fault is a structured transformation that keeps the information in
the weights; and an edit that makes things worse would be costly. Under these conditions the agent found
causes no script had been written for and never damaged an intact model.

\emph{Do not rely on an agent's ``no fault found''} when the model has no such reference and the loss is
within the sampling noise of the labelled pool: this is where all twelve misses occurred. Do not expect it
to recover information that is gone (reset statistics), to solve a large assignment problem reliably
(random permutations), or to keep looking after the first fault is repaired.

\emph{Do not use generic adaptation as a repair.} Re-estimating statistics, entropy minimisation and
fine-tuning recover part of some faults, but they also lower the score of intact models and of models
whose fault they do not match, and they do not say what was wrong.

\section{Limitations}
\label{sec:limits}
Each instance was run once and each condition has three instances, so the per-condition counts are
descriptive; repeated attempts on the same instances in the ablation show that single runs differ. One agent
model was used. The faults are synthetic, single or double, and of clean, structured kinds, including the
five ``unseen'' types; real pipelines fail in messier ways, and nothing here shows that the agent copes with
unstructured damage. The unseen-fault test is one-sided by construction (\cref{sec:unseen}), and the
post-hoc script covers segmentation and detection only. The setting excludes the most common remedy in
practice, comparison with the original model. The non-public model is a fine-tuned and permuted version of
a public one, so general priors about such networks remain available to the agent. Both judges of the
diagnoses are language models. Classification uses ImageNetV2 in place of the ImageNet validation set. The
isolation of the agents rests on rules and an audit of the transcripts, not on a technical barrier; six
runs were flagged or fell into grey areas of the rules (\cref{app:audit}), and without them 120 of 158
faulted models are repaired. The explanation of the detection failures by the mismatched statistics rests
on the agents' accounts and on the effect of re-estimation; it was not tested by an intervention as in the
ViT ablation. We did not measure cost: an agent run takes minutes to an hour of model calls, a script
seconds to minutes.

\section{Conclusion}
Weight editing by an agent is neither a new kind of repair nor an illusion. For deployment faults of vision
models, the edit itself is easy and the diagnosis can be scripted once the kinds of fault are known; a
few hundred lines do about as well as the agent. What the agent adds is that nobody has to know the kinds
in advance: it found and exactly inverted faults that no script had been written for, and it left intact
models alone. That advantage holds where the model carries a trustworthy reference for its own activations
and the fault is an invertible structured transformation, and it disappears where the reference is missing,
where the inverse is a search problem, and where one fault hides another. Claims about agents that edit
weights should be read against a script built for the same job; we hope the comparison becomes routine.

\section*{AI Use Statement}
The object of this study is a language-model agent
(Claude Opus 5.5, run as isolated sub-agents); its use is described in \cref{sec:setup}. An AI coding
assistant (Claude Code) was also used to implement the fault construction, tools, controls and scorers, to
run the experiments, to keep the experiment log, to carry out the transcript audit and the judgement of the
diagnoses under written criteria, and to draft this manuscript. The authors specified the research question,
the constraints of the protocol and the decisions between rounds, and are responsible for all content.

\section*{Impact Statement}
This paper studies when a language-model agent can find and undo faults in deployed vision models. The
intended use is maintenance: restoring a model to its released behaviour. The same capability, editing
weights from forward passes alone, could be used to alter models; our setting does not make this easier
than existing editing methods, since every edit is scored against a known reference.

\bibliographystyle{icml2027}

\newpage
\appendix
\onecolumn

\section{Fault Definitions and Informed Edits}
\label{app:faults}
All faults act on a public checkpoint. Random choices use a generator seeded by the instance seed.

\begin{itemize}
  \item \fault{bgr}, \fault{invert}, \fault{rot90}: applied to the image before the model's own
    preprocessing. Informed edit: channel order of the first layer exchanged (\fault{bgr}); the affine map
    folded into the first convolution and the first normalisation statistics (\fault{invert}; inexact at
    image borders for convolutional models); all kernels and, for ViT, the positional grid rotated
    (\fault{rot90}; inexact for ResNet-50 because of stride alignment, recovery 0.93).
  \item \fault{bnreset}: running mean and variance of a random 30\% of the BatchNorm layers set to 0 and 1.
    Informed: original buffers.
  \item \fault{rownoise}: eight output rows of one randomly chosen weight tensor overwritten with Gaussian
    noise whose standard deviation is $\lambda$ times that of the tensor,
    $\lambda\in\{2,5,20,200\}$ by the strength rule. Informed: original rows.
  \item \fault{perm}: the input channels of one $3\times3$ convolution in the second or third stage are
    permuted (ViT: the input columns of the second MLP layer of one block). \fault{perms}: a fraction of the
    channels, chosen so that the pool score falls by at least 3\% relative. Informed: inverse permutation.
  \item \fault{kflip}: kernels flipped in both spatial axes; scope by the strength rule from one
    convolution, all $3\times3$ convolutions of the third stage, of stages two to four, or every kernel
    larger than one (ViT: the patch-embedding kernel). Informed: flipped back.
  \item \fault{layout}: the weight of $n\in\{1,2,4\}$ convolutions stored as a permutation of its axes and
    reread in the original shape (ViT: the attention output projection transposed). Informed: inverse.
  \item \fault{swap}: the weights and buffers of two adjacent blocks of equal shape exchanged. Informed:
    exchanged back.
  \item \fault{clsshift}: rows of the final classifier shifted cyclically by one. \fault{boxconv}: for each
    class, the four box-regression outputs reordered from $(x,y,w,h)$ to $(y,x,h,w)$. \fault{scorecal}: a
    constant $s\in\{2,4,6,8\}$ added to the background bias of the box classifier ($s=6$ on all seeds).
    Informed: exact inverses.
  \item \fault{comboA}: \fault{invert} and \fault{layout} with $n=1$. \fault{comboB}: \fault{perm} and
    \fault{kflip} of one convolution in the same model. \fault{kflipr}: \fault{kflip} of the six
    $3\times3$ convolutions of the third stage, after which all BatchNorm statistics are re-estimated on the
    200 unlabelled pool images; informed: kernels flipped back and original statistics restored.
  \item Fault types outside every script library (\cref{sec:unseen}). \fault{sign}: the weight of one
    $3\times3$ convolution multiplied by $-1$. \fault{kroll}: the $3\times3$ kernels of the six convolutions
    of the third stage rolled circularly by one position in both axes. \fault{crev}: the input channels of
    one $3\times3$ convolution in reversed order. \fault{pairswap}: three disjoint random pairs of object
    classes exchanged in the weight and bias rows of the final classifier. \fault{bnswap}: weight and bias
    of one or two backbone BatchNorm layers exchanged. Informed: exact inverses.
\end{itemize}
\begin{table}[h]
\caption{Fault types per agent round (rounds as in \cref{tab:runs}). Every listed condition has three seeds
per model. Types that failed the validity criterion on a task were not given to agents in the first round of
that task; \fault{bgr}, \fault{perm} and \fault{swap} lose only 1 to 8\% relative on DeepLabV3 and were used
in the second segmentation round as small-drop faults.}
\label{tab:rounds}
\centering\small
\begin{tabular}{@{}lll@{}}
\toprule
Round & Models & Conditions \\
\midrule
C1 & ResNet-50, ViT-B/16 & \fault{bgr}, \fault{rownoise}, \fault{invert}, \fault{rot90}, \fault{kflip}, \fault{layout}, \fault{perm} \\
C2 & ResNet-50, ViT-B/16 & fault-free, \fault{swap}, \fault{perms}, \fault{comboA}, \fault{comboB} \\
C3 & ViT-B/16, two arms & fault-free, \fault{swap}, \fault{perms}, \fault{comboA}, \fault{comboB} \\
S1 & DeepLabV3-R50 & fault-free, \fault{invert}, \fault{kflip}, \fault{layout}, \fault{rownoise}, \fault{clsshift} \\
S2 & DeepLabV3-R50 & \fault{bgr}, \fault{perm}, \fault{swap}, \fault{comboA}, \fault{comboB}, \fault{kflipr} \\
D & Faster R-CNN R50-FPN & fault-free, \fault{bnreset}, \fault{perm}, \fault{kflip}, \fault{boxconv}, \fault{scorecal} \\
S3 & DeepLabV3-R50, private weights & fault-free, \fault{kflip}, \fault{layout}, \fault{kflipr} \\
S4 & DeepLabV3-R50 & \fault{sign}, \fault{kroll}, \fault{crev}, \fault{pairswap}, \fault{bnswap} \\
\bottomrule
\end{tabular}
\end{table}

\section{Task Text and Tool Interface}
\label{app:task}
The task text of the segmentation rounds follows verbatim. The classification text of rounds C2 and C3 and
the detection text differ only in the description of the model, the data and the metric; in C3 the arm with
a reference additionally describes the file of reference statistics. The text of round C1 stated that a
fault exists. Each agent was started with a fixed prompt that names its folder, tells it to carry out the
task in \texttt{TASK.md} without asking questions, to work only inside that folder, not to use the internet
or sub-agents, and to reply with a summary of at most 150 words.

{\fontsize{6.3}{7.4}\selectfont
\begin{Verbatim}
# Task: check a semantic segmentation model, and repair it if it is faulty

You are working alone in this folder. Work only inside it.

## Situation

`model.pt` holds the weights of a **DeepLabV3 with a ResNet-50 backbone** (torchvision architecture
`deeplabv3_resnet50`, 21 classes: background and the 20 Pascal VOC object classes, in the usual VOC order). It
labels every pixel of a colour image. On its original validation data this model was reported to reach about
78% mIoU. It has been moved into a new system. **We do not know whether anything is wrong.** There may be no
fault at all, one fault, or more than one. A fault may lie in the images the model receives, in the stored
weights, or in how its outputs are read, and it may cost many points of mIoU or only a few.

Your job:

1. **Diagnose.** Decide whether something is wrong. If so, state each cause and the evidence for it. If you find
   nothing, say so.
2. **Repair**, if there is something to repair, by editing the weights, so that the model does well on the
   deployment data as it arrives. The image pipeline itself cannot be changed; only the weights can.

We will score the model with your edit on a large, separate set of deployment images that you cannot see, by
mIoU (defined in the header of `forward.py`). An edit to a model that had no fault can only lose mIoU, and a
fault you miss stays in.

## What you have

- `model.pt` -- the state dict. You may load it and read the weights directly (for example with
  `torch.load("model.pt", map_location="cpu")`). Reading weights is free.
- `pool.pt` -- a small pool of deployment images exactly as the model receives them (already resized by the
  pipeline, shorter side 520 pixels; sizes differ between images): `xl` (200 images) with label maps `yl`
  (values 0..20, 255 = ignore), and `xu` (200 images) without labels. Lists of uint8 tensors. You may read them
  directly.
- `forward.py` -- **the only way you may run the model.** Read its header or run `python forward.py --info`
  (free). It returns the predicted label maps and, on request, the logits and the output of any module; on
  labelled pool images it also reports the mIoU.
- `config.json` -- the budget.

Run Python as `python -I <script>`.

## Rules

- The model may be run only through `forward.py`. Do not write your own forward pass of this network or of part
  of it, and do not instantiate the architecture yourself to run it. Arithmetic on weights and on arrays returned
  by `forward.py` is fine.
- Budget: at most **50 forward calls**, each on at most **200 images**. The tool refuses further
  calls. Refused calls are not counted. Plan before you spend.
- A forward call may use pool images or images you construct yourself (values in [0, 1]; see the tool header for
  the allowed sizes).
- An edit is an additive delta: an `.npz` file whose keys are state-dict names and whose arrays are added to the
  stored weights. Any entry may be edited, in any layer. You can test a delta with `forward.py --delta`.
- **No reference weights.** Do not load, download or look for any other copy of this architecture's weights
  (pretrained checkpoints, torchvision or other model hubs, caches or files elsewhere on this machine), and do
  not use any image data other than `pool.pt`. The diagnosis has to come from this model and this pool.
- No gradients of any kind: no backpropagation and no gradient-based optimiser (SGD, Adam and the like) anywhere,
  including in offline computations on weights, and no fitting of weights to the labels by iterative
  optimisation or by a search driven by repeated forward calls. Allowed: closed-form computations (means,
  covariances, least squares, matrix inverses, sorting), and gradient-free combinatorial solvers that work on
  arrays you already have (assignment or matching problems), as long as they do not call `forward.py` in a loop.
- If something you compute or choose uses the labels `yl`, say so and say how many labelled images it used. This
  includes choosing between candidate edits by their mIoU on the labelled pool.
- Do not modify `model.pt`, `pool.pt`, `forward.py`, `config.json` or `ledger.jsonl`.
- No internet, no sub-agents, no reading or writing files outside this folder.

## What to hand in

- `submission/diagnosis.md` -- at most 300 words. First line: `VERDICT: no fault found` or
  `VERDICT: fault(s) found`. Then what is wrong (each cause separately), the evidence, how confident you are, and
  what you could not determine. If you are unsure between explanations, say so; do not overstate.
- `submission/edit.npz` -- your final delta (use `np.savez_compressed`; include only entries you change). If you
  conclude that nothing is wrong, or that no edit helps, submit an empty `.npz` and say so.
- `submission/rebuild.py` -- a script that recreates `submission/edit.npz` from `model.pt`, `pool.pt` and files
  produced by your forward calls, without calling `forward.py` again.
- `notes.md` -- a short log of what you tried, in order, including what did not work, and the mIoU on the
  labelled pool you measured for your final edit (state which images it was measured on).

Report numbers as you measured them. "Nothing is wrong" and "I could not fix this, here is what I found" are
both valid results when they are true.
\end{Verbatim}
}

\noindent The header of the forward tool for segmentation, which the agent could read:

{\fontsize{6.3}{7.4}\selectfont
\begin{Verbatim}
forward.py -- the only permitted way to run the model in this folder.

  python forward.py --info
        free; lists module names, parameter names and shapes, pool layout, calls used / remaining.

  python forward.py --idx 0:200 --out out.npz [--delta edit.npz] [--logits] [--layers backbone.layer1,classifier.1] [--reduce pool]
        one counted forward call on pool images. Pool indices 0..199 are labelled, 200..399 unlabelled.
        --idx accepts "a:b", a comma list "3,7,9", or a .npy file of integer indices.

  python forward.py --input my_images.npy --out out.npz [...]
        one counted forward call on your own images: float array (N, 3, H, W) with values in [0, 1]; all images of
        one file have the same size; each side at least 64 pixels and H * W at most 800,000.

Each call takes at most `max_images` images (see config.json) and the number of calls is capped; a refused call
is not counted. The model runs in eval mode, float32, one image per forward pass, at the size the image has
(pool images are already at the pipeline size: shorter side 520). Images in [0, 1] are normalised inside this
tool with the fixed ImageNet mean (0.485, 0.456, 0.406) and std (0.229, 0.224, 0.225) before they reach the
network; this step is part of the pipeline and cannot be edited. The prediction is the argmax over the 21 class
channels of the model output "out" at the image size. The auxiliary head (aux_classifier.*) is in the state dict
but its output is not used by the pipeline.

--delta   .npz whose keys are state-dict names (see --info) and whose arrays have the same shape as the entry;
          they are ADDED to the stored weights for this call only. BatchNorm buffers (running_mean, running_var)
          can be edited the same way.
--logits  also return the class logits at the output stride (the output of module "classifier", before it is
          upsampled to the image size).
--layers  comma-separated module names; the output of each module is returned.
--reduce  how module outputs are returned:
            pool       (N, C)  mean per image over positions                      [default]
            chanstats  (2, C)  per-channel mean and variance over images and positions
            full       the whole output of each image as float16; refused above 2e8 values per call

out.npz (compressed) contains:
  idx                   pool indices, or -1 for your own images
  pred_<k>              uint8 (H, W) predicted class map of the k-th image of the call (k = 0, 1, ...)
  logits_<k>            float16 (21, h, w), with --logits
  act/<module>          with --reduce pool or chanstats
  act/<module>/<k>      float16 (C, h, w), with --reduce full
  conf, miou            only when every image of the call is a labelled pool image: the 21 x 21 confusion matrix
                        (rows = label, columns = prediction, pixels labelled 255 ignored) summed over the images,
                        and the mIoU = mean over the classes present in these labels of TP / (label + predicted - TP).
                        This is the metric the model will be scored with. The mIoU is also printed.
\end{Verbatim}
}

\noindent The detection tool returns boxes, scores and labels per image instead of label maps, and reports
COCO box AP on labelled pool images; the classification tool returns logits and accuracy.

\section{Script Library and Control Settings}
\label{app:scripts}
\begin{itemize}
  \item F1: re-estimation of all BatchNorm running statistics on unlabelled pool images.
  \item F2: ridge refit of the final classifier on the labelled pool; F3: bias correction of the final
    layer (classification and segmentation).
  \item F6: best of the six input channel orders on the labelled pool, written into the first layer.
  \item F7: per-channel affine standardisation of the input, folded into the first layer.
  \item F8: best cyclic class offset in $\{-2,\dots,2\}$.
  \item F9: in every convolution and linear layer, rows whose log-norm has a robust $z$-score above 6 are
    set to zero (at most 5\% of the rows of a tensor; prediction layers excluded where this harmed the
    fault-free model), followed by F1 except in detection.
  \item Try-all: no script, each script, each ordered pair; scripts fitted on one half of the labelled pool,
    candidates scored on the other half, the winner refitted on all labelled images.
  \item Entropy minimisation on normalisation parameters over the unlabelled pool.
  \item Fine-tuning on the labelled pool, last layer or all weights. Classification: 10 epochs, SGD,
    learning rate $10^{-3}$, momentum 0.9, batch 32. Detection: 5 epochs, SGD, learning rate
    $2.5\times10^{-3}$, momentum 0.9, weight decay $10^{-4}$, batch 2, horizontal flips.
    Segmentation: 10 epochs, SGD, learning rate $10^{-3}$, momentum 0.9, batches of 8 crops of
    $480\times480$. Entropy minimisation: one pass over the unlabelled pool, SGD, learning rate $10^{-3}$,
    momentum 0.9, batch 64 (classification) or 8 (segmentation); it was not run for detection, where
    re-estimation of the statistics (F1) is the test-time-adaptation control.
\end{itemize}

\paragraph{The post-hoc script LH (\cref{sec:lh}).}
Locate: forward hooks on the input of every BatchNorm layer, 50 unlabelled images, $d_\ell$ of \cref{eq:dev};
layers whose input has a single value per channel are left out. Suspects: the convolutions in front of the
three layers with the largest $d_\ell$, and the first layer in forward order with $d_\ell$ above twice the
median. Hypotheses: the five non-identity input channel orders; inversion folded into the first layer; class
shifts by $-2,\dots,2$ (segmentation); three alternative box-coordinate orders and the bias-sum correction
(detection); per suspect $3\times3$ convolution a flip in both axes, in each axis, a transposition and the
inverse axis order; per suspect block the exchange with the next block of equal shape, the residual branch
switched off, and re-estimated statistics; all $3\times3$ kernels of stage 2, 3 or 4 flipped, each alone and
followed by F1; F1; F9. Selection: every hypothesis is scored on 40 labelled images, the five best on the
first 100, and the best is accepted if it gains at least 0.01 (mIoU or AP) on the first and on the second
100 labelled images; at most three rounds.

\section{Per-Run Results}
\begin{longtable}{@{}llllrrlllr@{}}
\caption{All 183 agent runs. Rounds: C1 classification with a fault known to exist, C2 classification blind, C3 ViT reference ablation (REF: reference statistics supplied; REP: repeat without), S1 and S2 segmentation, D detection, S3 segmentation on non-public weights, S4 segmentation with fault types outside every script library. Recovery is computed by us on the locked test split; for fault-free models the column says whether the agent submitted a non-empty edit. Audit: runs flagged or marked grey in the rule audit.}\label{tab:runs}\\
\toprule
Round & Case & Model & Fault & Seed & Calls & Verdict & Diagnosis & Recovery & Audit \\
\midrule\endfirsthead
\toprule
Round & Case & Model & Fault & Seed & Calls & Verdict & Diagnosis & Recovery & Audit \\
\midrule\endhead
\bottomrule\endfoot
C1 & 5db3 & R50 & BGR & 0 & 3 & -- & correct & 1.00 &  \\
C1 & 2769 & R50 & BGR & 1 & 4 & -- & correct & 1.00 &  \\
C1 & 1e15 & R50 & BGR & 2 & 7 & -- & correct & 1.00 &  \\
C1 & fa30 & ViT & BGR & 0 & 4 & -- & correct & 1.00 &  \\
C1 & b2ff & ViT & BGR & 1 & 4 & -- & correct & 1.00 &  \\
C1 & 6f9c & ViT & BGR & 2 & 3 & -- & correct & 1.00 &  \\
C1 & 9429 & R50 & INVERT & 0 & 4 & -- & correct & 1.00 &  \\
C1 & 114c & R50 & INVERT & 1 & 9 & -- & correct & 1.00 &  \\
C1 & 2213 & R50 & INVERT & 2 & 4 & -- & correct & 1.00 &  \\
C1 & 74a9 & ViT & INVERT & 0 & 10 & -- & correct & 1.00 &  \\
C1 & 07ee & ViT & INVERT & 1 & 5 & -- & correct & 1.00 &  \\
C1 & 107b & ViT & INVERT & 2 & 2 & -- & correct & 1.00 &  \\
C1 & 2309 & R50 & ROT90 & 0 & 7 & -- & correct & 0.95 &  \\
C1 & 6082 & R50 & ROT90 & 1 & 5 & -- & correct & 0.94 &  \\
C1 & 39f1 & R50 & ROT90 & 2 & 7 & -- & correct & 0.96 &  \\
C1 & 523f & ViT & ROT90 & 0 & 7 & -- & correct & 1.00 &  \\
C1 & 7c0c & ViT & ROT90 & 1 & 7 & -- & correct & 1.00 &  \\
C1 & f071 & ViT & ROT90 & 2 & 5 & -- & correct & 1.00 &  \\
C1 & 67cf & R50 & ROWNOISE & 0 & 4 & -- & correct & 1.00 &  \\
C1 & cc90 & R50 & ROWNOISE & 1 & 5 & -- & correct & 1.00 &  \\
C1 & 906c & R50 & ROWNOISE & 2 & 3 & -- & correct & 1.00 &  \\
C1 & fe9a & ViT & ROWNOISE & 0 & 2 & -- & correct & 1.00 &  \\
C1 & 234d & ViT & ROWNOISE & 1 & 4 & -- & correct & 1.00 &  \\
C1 & 53af & ViT & ROWNOISE & 2 & 3 & -- & correct & 1.00 &  \\
C1 & 6f10 & R50 & PERM & 0 & 8 & -- & correct & 1.00 & grey \\
C1 & d5c0 & R50 & PERM & 1 & 10 & -- & correct & 1.00 &  \\
C1 & a5b6 & R50 & PERM & 2 & 12 & -- & correct & 1.00 &  \\
C1 & 440d & ViT & PERM & 0 & 15 & -- & wrong & 0.00 &  \\
C1 & 63dd & ViT & PERM & 1 & 23 & -- & correct & 0.86 &  \\
C1 & ac98 & ViT & PERM & 2 & 20 & -- & correct & 0.91 &  \\
C1 & 677d & R50 & KFLIP & 0 & 20 & -- & correct & 1.00 &  \\
C1 & 9a30 & R50 & KFLIP & 1 & 10 & -- & correct & 1.00 &  \\
C1 & 6a37 & R50 & KFLIP & 2 & 12 & -- & correct & 1.00 &  \\
C1 & d105 & ViT & KFLIP & 0 & 12 & -- & correct & 1.00 &  \\
C1 & b362 & ViT & KFLIP & 1 & 24 & -- & correct & 1.00 &  \\
C1 & 58d5 & ViT & KFLIP & 2 & 36 & -- & correct & 1.00 &  \\
C1 & 9439 & R50 & LAYOUT & 0 & 5 & -- & correct & 1.00 &  \\
C1 & 5b8a & R50 & LAYOUT & 1 & 16 & -- & partly & 0.85 &  \\
C1 & 5f3d & R50 & LAYOUT & 2 & 34 & -- & partly & 0.93 &  \\
C1 & 2a37 & ViT & LAYOUT & 0 & 3 & -- & correct & 1.00 &  \\
C1 & 563c & ViT & LAYOUT & 1 & 11 & -- & correct & 1.00 &  \\
C1 & 8cd2 & ViT & LAYOUT & 2 & 10 & -- & correct & 1.00 &  \\
\midrule
C2 & 8f4b & R50 & NONE & 0 & 3 & no fault & correct & no edit &  \\
C2 & fe4c & R50 & NONE & 1 & 4 & no fault & correct & no edit &  \\
C2 & 60e5 & R50 & NONE & 2 & 6 & no fault & correct & no edit &  \\
C2 & 5db3 & ViT & NONE & 0 & 6 & no fault & correct & no edit & flagged \\
C2 & c244 & ViT & NONE & 1 & 6 & no fault & correct & no edit &  \\
C2 & 942a & ViT & NONE & 2 & 6 & no fault & correct & no edit &  \\
C2 & 944d & R50 & PERMS & 0 & 15 & fault & correct & 0.92 &  \\
C2 & 9c6e & R50 & PERMS & 1 & 7 & fault & partly & 0.37 &  \\
C2 & db4f & R50 & PERMS & 2 & 6 & fault & partly & 0.47 &  \\
C2 & 9605 & ViT & PERMS & 0 & 14 & no fault & wrong & 0.00 &  \\
C2 & 0eb4 & ViT & PERMS & 1 & 16 & no fault & wrong & 0.00 &  \\
C2 & a023 & ViT & PERMS & 2 & 5 & no fault & wrong & 0.00 &  \\
C2 & 4680 & R50 & SWAP & 0 & 13 & fault & correct & 1.00 &  \\
C2 & a55a & R50 & SWAP & 1 & 7 & fault & correct & 0.98 &  \\
C2 & e188 & R50 & SWAP & 2 & 12 & fault & correct & 1.01 &  \\
C2 & 6402 & ViT & SWAP & 0 & 10 & no fault & wrong & 0.00 &  \\
C2 & 9281 & ViT & SWAP & 1 & 3 & fault & correct & 1.00 &  \\
C2 & bbfd & ViT & SWAP & 2 & 23 & fault & correct & 1.00 &  \\
C2 & 2650 & R50 & COMBOA & 0 & 28 & fault & partly & 0.97 &  \\
C2 & d955 & R50 & COMBOA & 1 & 7 & fault & correct & 1.00 &  \\
C2 & b749 & R50 & COMBOA & 2 & 14 & fault & correct & 1.00 &  \\
C2 & b4c8 & ViT & COMBOA & 0 & 6 & fault & partly & 0.92 &  \\
C2 & 3347 & ViT & COMBOA & 1 & 4 & fault & partly & 0.94 & flagged \\
C2 & 22df & ViT & COMBOA & 2 & 6 & fault & partly & 0.92 &  \\
C2 & b4bd & R50 & COMBOB & 0 & 10 & fault & correct & 1.00 &  \\
C2 & e665 & R50 & COMBOB & 1 & 7 & fault & correct & 0.99 &  \\
C2 & 8b90 & R50 & COMBOB & 2 & 22 & fault & partly & 0.57 &  \\
C2 & 5b57 & ViT & COMBOB & 0 & 26 & fault & partly & 0.73 &  \\
C2 & ea0d & ViT & COMBOB & 1 & 40 & fault & partly & 0.30 &  \\
C2 & df32 & ViT & COMBOB & 2 & 35 & fault & partly & 0.45 &  \\
\midrule
C3-REF & 8ec4 & ViT & NONE & 0 & 2 & no fault & correct & no edit &  \\
C3-REF & c2d1 & ViT & NONE & 1 & 3 & no fault & correct & no edit &  \\
C3-REF & d407 & ViT & NONE & 2 & 2 & no fault & correct & no edit &  \\
C3-REP & ca9e & ViT & NONE & 0 & 8 & no fault & correct & no edit &  \\
C3-REP & ea44 & ViT & NONE & 1 & 14 & no fault & correct & no edit &  \\
C3-REP & 32d2 & ViT & NONE & 2 & 11 & no fault & correct & no edit &  \\
C3-REF & 9edc & ViT & PERMS & 0 & 7 & fault & correct & 0.91 &  \\
C3-REF & 737f & ViT & PERMS & 1 & 6 & fault & correct & 0.68 &  \\
C3-REF & d00c & ViT & PERMS & 2 & 8 & fault & correct & 0.85 &  \\
C3-REP & 5dda & ViT & PERMS & 0 & 11 & no fault & wrong & 0.00 &  \\
C3-REP & 119a & ViT & PERMS & 1 & 11 & no fault & wrong & 0.00 &  \\
C3-REP & e8da & ViT & PERMS & 2 & 13 & fault & correct & 0.68 &  \\
C3-REF & 0d4c & ViT & SWAP & 0 & 3 & fault & correct & 1.00 &  \\
C3-REF & 017f & ViT & SWAP & 1 & 3 & fault & correct & 1.00 &  \\
C3-REF & d1f4 & ViT & SWAP & 2 & 4 & fault & correct & 1.00 &  \\
C3-REP & 3f12 & ViT & SWAP & 0 & 12 & no fault & wrong & 0.00 &  \\
C3-REP & a1db & ViT & SWAP & 1 & 11 & no fault & wrong & 0.00 &  \\
C3-REP & 7c45 & ViT & SWAP & 2 & 31 & fault & correct & 1.00 &  \\
C3-REF & 4b6f & ViT & COMBOA & 0 & 6 & fault & correct & 1.00 &  \\
C3-REF & 609d & ViT & COMBOA & 1 & 6 & fault & correct & 1.00 &  \\
C3-REF & 0f8c & ViT & COMBOA & 2 & 4 & fault & correct & 1.00 &  \\
C3-REP & 9c74 & ViT & COMBOA & 0 & 6 & fault & partly & 0.92 &  \\
C3-REP & 4578 & ViT & COMBOA & 1 & 10 & fault & correct & 1.00 &  \\
C3-REP & 701b & ViT & COMBOA & 2 & 5 & fault & correct & 1.00 &  \\
C3-REF & d9df & ViT & COMBOB & 0 & 9 & fault & partly & 0.42 &  \\
C3-REF & 57e4 & ViT & COMBOB & 1 & 12 & fault & partly & 0.32 &  \\
C3-REF & 444c & ViT & COMBOB & 2 & 12 & fault & partly & 0.43 &  \\
C3-REP & efa1 & ViT & COMBOB & 0 & 25 & fault & partly & 0.42 &  \\
C3-REP & 22d5 & ViT & COMBOB & 1 & 23 & fault & partly & 0.34 &  \\
C3-REP & 9ec4 & ViT & COMBOB & 2 & 22 & fault & partly & 0.45 &  \\
\midrule
S1 & de3e & DLv3 & NONE & 0 & 5 & no fault & correct & no edit &  \\
S1 & 2e25 & DLv3 & NONE & 1 & 4 & no fault & correct & no edit &  \\
S1 & f746 & DLv3 & NONE & 2 & 6 & no fault & correct & no edit & flagged \\
S1 & c294 & DLv3 & INVERT & 0 & 13 & fault & correct & 1.00 &  \\
S1 & 76bd & DLv3 & INVERT & 1 & 5 & fault & correct & 1.00 &  \\
S1 & 16ad & DLv3 & INVERT & 2 & 17 & fault & correct & 1.00 &  \\
S1 & 8c04 & DLv3 & ROWNOISE & 0 & 2 & fault & correct & 1.00 &  \\
S1 & ff64 & DLv3 & ROWNOISE & 1 & 3 & fault & correct & 0.99 &  \\
S1 & 9a95 & DLv3 & ROWNOISE & 2 & 5 & fault & correct & 1.00 &  \\
S1 & c2cc & DLv3 & KFLIP & 0 & 9 & fault & correct & 1.00 &  \\
S1 & 0cf6 & DLv3 & KFLIP & 1 & 9 & fault & correct & 1.00 &  \\
S1 & c6af & DLv3 & KFLIP & 2 & 13 & fault & correct & 1.00 &  \\
S1 & a300 & DLv3 & LAYOUT & 0 & 6 & fault & correct & 1.00 &  \\
S1 & 9acc & DLv3 & LAYOUT & 1 & 13 & fault & correct & 1.00 &  \\
S1 & b7a5 & DLv3 & LAYOUT & 2 & 4 & fault & correct & 1.00 &  \\
S1 & 2b88 & DLv3 & CLSSHIFT & 0 & 3 & fault & correct & 1.00 &  \\
S1 & 124d & DLv3 & CLSSHIFT & 1 & 3 & fault & correct & 1.00 &  \\
S1 & e62b & DLv3 & CLSSHIFT & 2 & 4 & fault & correct & 1.00 &  \\
\midrule
S2 & b2c9 & DLv3 & BGR & 0 & 3 & fault & correct & 0.99 &  \\
S2 & a2a0 & DLv3 & BGR & 1 & 6 & fault & correct & 0.99 &  \\
S2 & 2159 & DLv3 & BGR & 2 & 5 & fault & correct & 1.01 &  \\
S2 & bd56 & DLv3 & PERM & 0 & 18 & fault & correct & 1.04 &  \\
S2 & 446b & DLv3 & PERM & 1 & 4 & fault & correct & 0.27 &  \\
S2 & db1b & DLv3 & PERM & 2 & 11 & fault & correct & 0.29 &  \\
S2 & 83e6 & DLv3 & KFLIPR & 0 & 47 & fault & correct & 0.98 &  \\
S2 & d58c & DLv3 & KFLIPR & 1 & 48 & fault & correct & 0.89 &  \\
S2 & 7a55 & DLv3 & KFLIPR & 2 & 47 & fault & correct & 0.99 &  \\
S2 & ce07 & DLv3 & SWAP & 0 & 25 & fault & partly & 0.43 &  \\
S2 & 3c9d & DLv3 & SWAP & 1 & 13 & fault & partly & -0.02 &  \\
S2 & 0757 & DLv3 & SWAP & 2 & 6 & fault & correct & 1.00 &  \\
S2 & 5ec2 & DLv3 & COMBOA & 0 & 15 & fault & partly & 0.94 &  \\
S2 & 4e1d & DLv3 & COMBOA & 1 & 8 & fault & correct & 1.00 &  \\
S2 & fd7b & DLv3 & COMBOA & 2 & 9 & fault & correct & 1.00 &  \\
S2 & 3692 & DLv3 & COMBOB & 0 & 13 & fault & partly & 0.52 &  \\
S2 & c7d7 & DLv3 & COMBOB & 1 & 9 & fault & partly & 0.77 &  \\
S2 & d84a & DLv3 & COMBOB & 2 & 8 & fault & partly & 0.70 &  \\
\midrule
D & 84eb & FRCNN & NONE & 0 & 5 & no fault & correct & no edit &  \\
D & e636 & FRCNN & NONE & 1 & 3 & no fault & correct & no edit &  \\
D & 44aa & FRCNN & NONE & 2 & 8 & no fault & correct & no edit &  \\
D & 5cab & FRCNN & BNRESET & 0 & 23 & fault & correct & 0.72 &  \\
D & c7de & FRCNN & BNRESET & 1 & 34 & fault & correct & 0.59 &  \\
D & f863 & FRCNN & BNRESET & 2 & 22 & fault & correct & 0.68 &  \\
D & 05fd & FRCNN & PERM & 0 & 17 & fault & correct & 0.08 &  \\
D & b3d7 & FRCNN & PERM & 1 & 8 & fault & correct & 1.00 &  \\
D & a79a & FRCNN & PERM & 2 & 3 & no fault & wrong & 0.00 &  \\
D & 5914 & FRCNN & KFLIP & 0 & 5 & no fault & wrong & 0.00 &  \\
D & 1a1a & FRCNN & KFLIP & 1 & 36 & fault & wrong & 0.00 & flagged \\
D & d274 & FRCNN & KFLIP & 2 & 3 & no fault & wrong & 0.00 &  \\
D & 0c7d & FRCNN & BOXCONV & 0 & 6 & fault & correct & 1.00 &  \\
D & 9a88 & FRCNN & BOXCONV & 1 & 6 & fault & correct & 1.00 &  \\
D & 3bed & FRCNN & BOXCONV & 2 & 4 & fault & correct & 1.00 &  \\
D & 7947 & FRCNN & SCORECAL & 0 & 5 & fault & correct & 1.00 &  \\
D & 26b5 & FRCNN & SCORECAL & 1 & 7 & fault & correct & 1.00 &  \\
D & 0886 & FRCNN & SCORECAL & 2 & 4 & fault & correct & 1.01 & flagged \\
\midrule
S3 & d1fb & DLv3 & NONE & 0 & 3 & no fault & correct & no edit &  \\
S3 & 4ec7 & DLv3 & NONE & 1 & 5 & no fault & correct & no edit &  \\
S3 & 9b55 & DLv3 & NONE & 2 & 6 & no fault & correct & no edit &  \\
S3 & 5d27 & DLv3 & KFLIP & 0 & 6 & fault & correct & 1.00 &  \\
S3 & 139e & DLv3 & KFLIP & 1 & 10 & fault & correct & 1.00 &  \\
S3 & 86df & DLv3 & KFLIP & 2 & 13 & fault & correct & 1.00 &  \\
S3 & a3ff & DLv3 & KFLIPR & 0 & 44 & fault & correct & 0.98 &  \\
S3 & 1ed6 & DLv3 & KFLIPR & 1 & 40 & fault & correct & 0.71 &  \\
S3 & 1e41 & DLv3 & KFLIPR & 2 & 45 & fault & correct & 0.98 &  \\
S3 & ae53 & DLv3 & LAYOUT & 0 & 7 & fault & correct & 1.00 &  \\
S3 & 52ab & DLv3 & LAYOUT & 1 & 6 & fault & correct & 1.00 &  \\
S3 & bec2 & DLv3 & LAYOUT & 2 & 4 & fault & correct & 1.00 &  \\
\midrule
S4 & fd8e & DLv3 & SIGN & 0 & 4 & fault & correct & 1.00 &  \\
S4 & 90a0 & DLv3 & SIGN & 1 & 7 & fault & correct & 1.00 &  \\
S4 & 3193 & DLv3 & SIGN & 2 & 3 & fault & correct & 1.00 &  \\
S4 & a9a6 & DLv3 & KROLL & 0 & 4 & fault & correct & 0.97 &  \\
S4 & ff17 & DLv3 & KROLL & 1 & 32 & fault & correct & 1.00 &  \\
S4 & a6e7 & DLv3 & KROLL & 2 & 6 & fault & correct & 1.00 &  \\
S4 & 2ce7 & DLv3 & CREV & 0 & 3 & fault & correct & 1.00 &  \\
S4 & c8c6 & DLv3 & CREV & 1 & 4 & fault & correct & 1.00 &  \\
S4 & 4a6d & DLv3 & CREV & 2 & 5 & fault & correct & 1.00 &  \\
S4 & 8267 & DLv3 & PAIRSWAP & 0 & 3 & fault & correct & 1.00 &  \\
S4 & fc4f & DLv3 & PAIRSWAP & 1 & 3 & fault & correct & 1.00 &  \\
S4 & 4227 & DLv3 & PAIRSWAP & 2 & 4 & fault & correct & 1.00 &  \\
S4 & 22fa & DLv3 & BNSWAP & 0 & 5 & fault & correct & 1.00 &  \\
S4 & 41fa & DLv3 & BNSWAP & 1 & 6 & fault & correct & 1.00 &  \\
S4 & df8d & DLv3 & BNSWAP & 2 & 6 & fault & correct & 1.00 &  \\
\end{longtable}

\section{Audit}
\label{app:audit}
After each round the transcript of every agent and the scripts left in its folder were scanned for:
network access and package installation; loading of pretrained or cached weights; paths outside the folder
of the agent; code that instantiates or re-implements the network or parts of it; gradient computation and
optimisers; and loops that call the forward tool. Every hit was read individually by the experimenter (the AI coding assistant, under the authors'
direction) and a verdict was written
into the log before the scores of that round were looked at. No run accessed the network, reference weights,
the answer keys or the folder of another agent. Six runs were flagged or marked grey:
\begin{itemize}
  \item C1, ResNet-50 \fault{perm} seed 0: the channel assignment was solved with a relaxation optimised by
    Adam on a closed-form statistic of one layer, without labels. The task text forbade backpropagation and
    allowed offline assignment solvers. Recovery 1.00.
  \item C2, two ViT runs (one fault-free, one \fault{comboA}): temporary files were written to a scratch
    directory outside the folder of the agent, to which the harness had pointed; the transcripts show no
    listing or reading of other files there. The directory was emptied before all later rounds.
  \item S1, one fault-free run: the final $1\times1$ layer was recomputed from returned features as a
    consistency check; the edit was empty.
  \item D, \fault{scorecal} seed 2 and \fault{kflip} seed 1: the parameter-free post-processing of the
    detector was re-implemented on outputs returned by the tool and used to choose a bias value by AP on the
    labelled pool. Recoveries 1.01 and 0.00.
\end{itemize}
Without these six runs, 120 of 158 faulted models are repaired (with them: 123 of 162); per round, C1 40 of 41,
C2 13 of 23, D 6 of 13, and the other rounds are unchanged. Rounds S3 and S4 had no flagged run. Level-by-level re-estimation of normalisation
statistics with one forward call per level (closed-form statistics, no labels) occurred in several runs and
was not counted as a search driven by forward calls. Several agents chose between a few discrete candidate
edits by the score on the labelled pool and declared it; this was allowed.

\section{Expectations Written Before Each Round}
\label{app:expect}
Before each round we wrote down what we expected, in counts. \Cref{tab:expect} lists the expectations that
turned out wrong; the remaining ones (false alarms and harm on fault-free models, in-library types within
0.1 of try-all, the ranges for the overall counts, and most per-type bounds) held. In eight of the
twenty-one rows the agents did better than predicted.

\begin{table}[h]
\caption{Expectations that were wrong. $+$: the agents did better than expected; $-$: worse (for the
scripts of the reference ablation: the scripts did worse); $\circ$: a control did better than expected.}
\label{tab:expect}
\centering\small
\begin{tabular}{@{}llll@{}}
\toprule
Round & Expectation & Outcome & \\
\midrule
C1 & \fault{rot90} on ResNet-50 repaired in at most 1 of 3 & 3 of 3 & $+$ \\
C1 & \fault{kflip} diagnosed and repaired in at most 2 of 6 & 6 of 6 & $+$ \\
C2 & \fault{swap} found in at most 3 of 6, repaired in at most 2 & 5 and 5 & $+$ \\
C2 & \fault{perms} repaired in at least 2 of 6 & 1 & $-$ \\
C2 & both components of \fault{comboA} found in at least 4 of 6 & 2 & $-$ \\
C2 & \fault{comboB}: both found and repaired in at least 3 of 6 & 2 and 2 & $-$ \\
C3 & without reference, both components in at most 1 of 6 compound runs & 2 & $+$ \\
C3 & with reference, both components in at least 4 of 6 compound runs & 3 & $-$ \\
C3 & with reference, \fault{comboB} repaired in at least 2 of 3; overall at least 9 of 12 & 0; 8 & $-$ \\
C3 scripts & closed-form alignment harmless on fault-free models & $-1.1$ to $-1.8$ points & $-$ \\
C3 scripts & reference detector locates at least 5 of 6 single faults & 4 & $-$ \\
S2 & \fault{perm} repaired in at least 2 of 3 & 1 & $-$ \\
S2 & exchange named for \fault{swap} in at least 2 of 3 & 1 & $-$ \\
S2 & \fault{comboB} repaired in at least 2 of 3 & 0 & $-$ \\
S2 & \fault{kflipr} diagnosed in 1 or 2 of 3, repaired in at most 1 & 3 and 3 & $+$ \\
D & \fault{kflip} diagnosed and repaired in at least 2 of 3 & 0 and 0 & $-$ \\
D & \fault{bnreset} repaired in 1 or 2 of 3 & 0 (recovery 0.59--0.72) & $-$ \\
D & \fault{scorecal} named and repaired in 1 or 2 of 3; at most 0.2 above fine-tuning & 3 and 3; $+0.33$ & $+$ \\
S4 & \fault{crev} and \fault{kroll} repaired in 1 or 2 of 3 each & 3 and 3 & $+$ \\
S4 & 9 to 12 of 15 repaired & 15 & $+$ \\
S4 & try-all at most 0.4 on every new type & \fault{sign} 0.84, \fault{pairswap} 0.69, \fault{bnswap} 0.64 & $\circ$ \\
\bottomrule
\end{tabular}
\end{table}

\end{document}